\documentclass[a4paper, fleqn, review, 11pt]{cas-sc}

\usepackage[numbers, sort&compress]{natbib}
\usepackage{threeparttable}
\usepackage{graphicx}
\usepackage{subcaption}

\begin{document}
\let\WriteBookmarks\relax
\def\floatpagepagefraction{1}
\def\textpagefraction{.001}
\let\printorcid\relax

\shorttitle{EdgeConvFormer}

\shortauthors{Jie Liu et~al.}

\title [mode = title]{EdgeConvFormer: Dynamic Graph CNN and Transformer based Anomaly Detection in Multivariate Time Series}                      




\author[1]{Jie Liu}






\affiliation[1]{organization={School of Electrical Engineering, Computing and Mathematical Sciences, Curtin University},
    city={Perth},
    postcode={WA 6102}, 
    country={Australia}}

\author[1]{Qilin Li}

\cormark[1]
\author[1]{Senjian An}


\author%
[1]
{Bradley Ezard}

\author%
[1]
{Ling Li}
\cormark[1]

\cortext[cor1]{Corresponding author\\
E-mail addresses: l.li@curtin.edu.au (L. Li), qilin.li@curtin.edu.au (Q. Li)}


\begin{abstract}
Transformer-based models for anomaly detection in multivariate time series can benefit from the self-attention mechanism due to its advantage in modeling long-term dependencies. However, Transformer-based anomaly detection models have problems such as a large amount of data being required for training, standard positional encoding is not suitable for multivariate time series data, and the interdependence between time series is not considered. To address these limitations, we propose a novel anomaly detection method, named EdgeConvFormer, which integrates Time2vec embedding, stacked dynamic graph CNN, and Transformer to extract global and local spatial-time information. This design of EdgeConvFormer empowers it with decomposition capacities for complex time series, progressive spatiotemporal correlation discovery between time series, and representation aggregation of multi-scale features. Experiments demonstrate that EdgeConvFormer can learn the spatial-temporal correlations from multivariate time series data and achieve better anomaly detection performance than the state-of-art approaches on many real-world datasets of different scales.

\end{abstract}




\maketitle

\section{Introduction}

In the modern manufacturing industry and engineering services, a large number of sensors are deployed to monitor the status and behaviors of many complex systems, generating large amounts of multivariate time series data. A critical task in managing such systems is to detect anomalies and ideally to localize the root cause of the anomalies so that the underlying issues can be resolved in a timely manner.

Time series data can be divided into univariate and multivariate time series. Univariate time series is the time series of a single variable without considering the influence of other variables. Multivariate time series are time series composed of multiple time-related variables where each variable not only depends on its past value but also on other variables. Complex systems are usually high-dimensional, of varying lengths, and interdependent among variables. The multiple variables in a multivariate time series often have a dynamic and cooperative relationship, reflecting the status of complex entities such as network services and large industrial equipment \cite{magan2020multivariate}.

Time series anomalies are usually defined in three scales: point, contextual, and collective. Point anomalies are considered global outliers referring to points that deviate significantly from the rest of the points. They are usually the spikes in the series. Contextual anomalies are local anomalies referring to data points that deviate from adjacent points within a certain range, such as the discord points within the same harmonic pattern. The first two types of anomalies focus on individual points. The third type, collective anomaly, refers to a sequence of points jointly forming an anomaly pattern. The individual points of a collective anomaly may not be anomalous by themselves, but co-occurrence of them becomes an anomaly \cite{lai2021revisiting, fisch2022real}. Identifying contextual and collective anomalies are more challenging. Ideally, an anomaly detector should be able to detect all three forms of anomalies.

The challenges of anomaly detection in time series generally include 1) the lack of specific definition of anomaly patterns; 2) noise in the input data; 3) increased computational complexity with the length of time series; 4) the difficulty to capture the interdependency of time series which are usually nonlinear and nonstationary; 5) difficult and expensive labeling of time series data.

Most of the existing anomaly detection methods for multivariate time series data are unsupervised, based on the ideas of reconstruction or future series prediction \cite{pang2021deep}. Reconstruction-based models minimize the reconstruction errors of training data (no anomalies), and the reconstruction errors of abnormal events are expected to be larger than those of normal ones. Hence the reconstruction-based models aim to enlarge the disparities of reconstruction errors between normal and abnormal data. On the other hand, future series prediction methods assume that normal events are predictable while abnormal ones are unpredictable. Robustness against noise is one of the main concerns of prediction-based models.

Compared to traditional statistical methods such as autoregressive moving-average (ARMA) \cite{pincombe2005anomaly}, and machine learning methods such as Local Outlier Factor (LOF)\cite{breunig2000lof}, Isolation Forest \cite{liu2008isolation}, One-Class Support Vector Machines (OC-SVM) \cite{chalapathy2018anomaly}, deep learning-based anomaly detection models \cite{zhou2017anomaly,zhang2019deep,su2019robust,meng2020localizing,ganokratanaa2020unsupervised} have made remarkable progress recently, due to their capability of learning hierarchical discriminative features from data without requiring domain experts to choose features manually. Among them the Transformer models \cite{xu2021anomaly, doshi2022tisat,tuli2022tranad}, based on the self-attention mechanism, have shown a significant advantage in modeling long-term dependencies for sequential data, enabling a more powerful representation and higher performance.

However, the current Transformer-based models have the following limitations: 1) The standard positional encoding is suboptimal for time series. The vanilla Transformer \cite{vaswani2017attention} uses cosine and sine functions to capture the position of words, and the existing Transformer-based models generally follow the same approach. Time series data is a special sequence whose temporal information reflects the order of timestamps and contains both periodic and aperiodic patterns. Since the frequencies and phase shifts of the sine function in the vanilla Transformer's position encoding are fixed, the same position encoding will be generated for timestamps at the corresponding positions in all sliding windows, thus ignoring the periodicity and intricate patterns of the temporal information. Hence an effective positional embedding method better suited for time series is worth studying. 2) Transformers lack inherent inductive biases, such as the translational invariance and locality of CNN. Therefore, they need massive amounts of data to train \cite{dosovitskiy2020image}. To alleviate this problem, the second generation of Vision Transformers (ViTs) mixed convolutional layers with attention layers to provide local inductive biases \cite{liu2021swin}. Transformer-based models for multivariate time series should also benefit from additional convolutional layers. 3) The existing Transformer models for multivariate time series anomaly detection do not consider the topology between sensors, which limits their ability to detect and explain anomalies occurring among multiple sensors. As abnormal events occur, it is usually the sensors' interrelationships that deviate from normal patterns. Therefore, exploring the inter-sensor correlation can provide more information for anomaly detection in multivariate time series.

\begin{figure}[ht]
\centering
\includegraphics[width=0.35\linewidth]{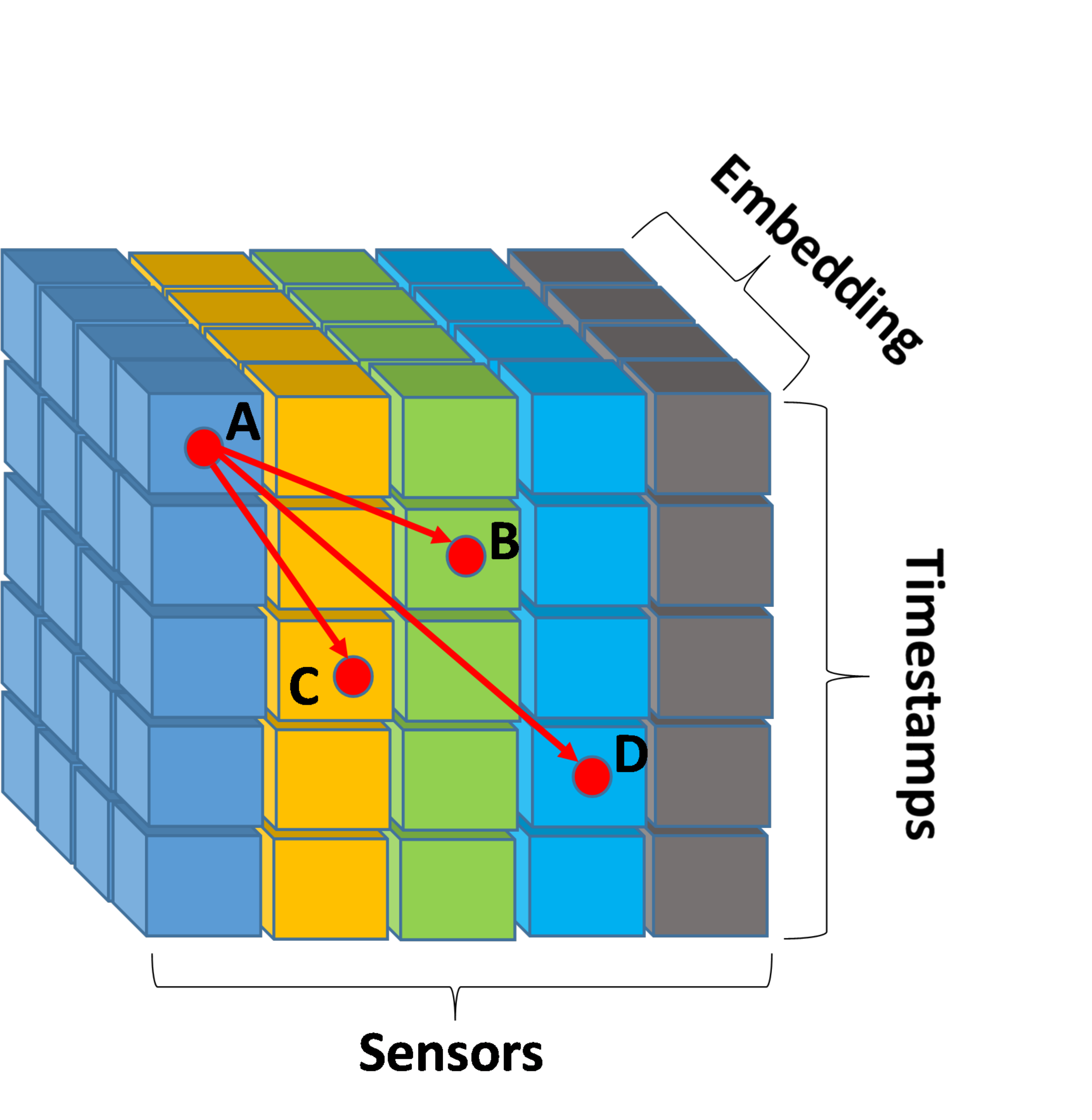}

\caption{The spatiotemporal topology}

\label{fig:fig1}
\end{figure}

Modeling the interdependencies between sensors and time steps using GNNs (Graph Neural Networks)\cite{defferrard2016convolutional} can enrich the representation of multivariate time series. \citeauthor{geng2022graph} \cite{geng2022graph} utilized GAT (Graph Attention Network) to extract multi-head temporal correlation across multiple time steps for the multivariate time series prediction task, however, it did not consider the spatial topology structure between sensors. To model the interdependencies between sensors (variables), GNNs typically require prior knowledge about the graph structure (i.e., stable interconnections between variables). For a complex system, it is difficult to derive complete topological information. \citeauthor{deng2021graph} \cite{deng2021graph} proposed graph structure learning by computing the similarity of embedding vectors between a sensor and all other sensors. The top k similarity values are chosen to construct the adjacent matrix for the graph structure, and graph attention is used to fuse a node's information with its neighbors based on the learned graph structure. This learned topological structure is the spatial topology which is a static topology. In reality, a sensor at one time instance may correlate to one or more sensors at another time instance, and such relationships are dynamic. For example, in a water plant, the failure of a valve will subsequently cause a chain reaction of the sensors in the pipeline associated with it. Abnormalities may not synchronously occur since there could be delays. Therefore, it would provide richer information and improve the accuracy of anomaly detection if the topological relationship can be extended from the spatial level to the spatiotemporal level. An example of a spatiotemporal topology is shown in Fig.~\ref{fig:fig1}, which is a three-dimensional space (sensors, timestamps, embedding). The embedding value of point A (a certain sensor at a certain time instant) is related to the embedding value of other space-time states B, C, and D.

To address the challenges identified above, we propose the following strategies in this paper. 1) For positional encodings, we use Time2vec \cite{mehran2019time2vec}, a learnable vector representation (or embedding) for time, to encode the multivariate time series by processing the time series data of each sensor individually. Time2vec is used to learn weights, frequencies, and phase shifts of sinusoidal functions from the data to capture the periodic behavior, while a linear function is used to capture the non-periodic pattern. As the sinusoidal functions in Time2Vec can represent continuous time instead of discrete positions and also capture periodic behaviors, which is not covered in the positional encoding of vanilla Transformers; 2) For the lack of locality of the Transformer and to find the spatiotemporal topological relationship between sensors, we deploy a multi-layer stack in which each layer consists of a dynamic graph CNN (dubbed EdgeConv in \cite{wang2019dynamic}) and a Transformer to refine the topological structure of the spatiotemporal state in the embedding space.  In each layer, EdgeConv is used to get the topological structure and edge features, the subsequent Transformer is used to attend to the time dimension for capturing the information across long-distance timestamps. The proposed model is hereby named EdgeConvFormer. The contributions of this paper are threefold:

\begin{enumerate}
\item Time2Vec is used to take advantage of the inherent properties of time series, i.e., the periodicity and intricate patterns of the temporal information.
\item EdgeConv is introduced into the multivariate time series anomaly detection to derive spatiotemporal level topological relationships between sensors, and make up for the transformer's lack of locality by searching the local graph for the most relevant points.
\item EdgeConv and Transformer are integrated in a hierarchical, multi-scale manner and reinforce each other at each layer to form an EdgeConvFormer model. The combination and refinement of multi-scale global and local features can improve the representation ability of embedding.
\end{enumerate}

Extensive experiments are conducted on publicly available datasets to evaluate and compare the proposed EdgeConvFormer against the state-of-the-art methods. EdgeConvFormer achieves the best or comparable anomaly detection performance under various evaluation metrics.

\section{Related work}
\subsection{Deep learning-based anomaly detection}
Deep learning-based anomaly detection techniques have greatly improved the anomaly detection performance compared to traditional statistical and machine learning methods \cite{pang2021deep, mehran2019time2vec}. The basic building blocks are CNN, RNN/LSTM, Autoencoder (AE), variants of AE, generative models, and deep one-class detection models. For example, \citeauthor{zhou2017anomaly} \cite{zhou2017anomaly} proposed Robust Deep AutoEncoder (RDA) for anomaly detection for cases with noisy data. \citeauthor{hundman2018detecting}\cite{hundman2018detecting} used LSTM to predict spacecraft telemetry and identify point outliers within each variable in a multivariate time series. 
\citeauthor{park2018multimodal}\cite{park2018multimodal} (LSTM-VAE) presented the LSTM-VAE model which employed the LSTM backbone for temporal modeling and the Variational AutoEncoder (VAE) for reconstruction to detect anomalies in robot-assisted feeding. \citeauthor{zhang2019deep}\cite{zhang2019deep}(MSCRED) used a combination of convolutional layers, LSTM layers, and an attention mechanism to construct an encoder-decoder structure to reconstruct the input signature matrices, i.e., matrices representing cross-correlation between channels, in which the residual signature matrices were utilized to pinpoint anomalies. \citeauthor{xu2018unsupervised}\cite{xu2018unsupervised} proposed an unsupervised anomaly detection algorithm based on VAE to deal with historical anomalies and missing data points. \citeauthor{zong2018deep}\cite{zong2018deep}(DAGMM) proposed an unsupervised multivariate anomaly detection method using the scoring function - a Gaussian Mixture Model and trained on an end-to-end AE framework. \citeauthor{su2019robust}\cite{su2019robust}(OmniAnomaly) suggested a more complex approach based on a VAE with Gated Recurrent Units (GRU) and used stochastic variable connection and planar normalizing flow to improve anomaly detection performance. \citeauthor{schmidt2019normalizing}\cite{schmidt2019normalizing} used a flow-based deep generative model to detect anomaly time series. \citeauthor{wen2019time}\cite{wen2019time} proposed a time series segmentation approach based on CNN and used transfer learning for anomaly detection. \citeauthor{zhou2019beatgan}\cite{zhou2019beatgan}(Beat-GAN) proposed an anomaly detection model, Beat-GAN, which detects anomalies using adversarially generated time series. \citeauthor{thill2020time}\cite{thill2020time}(TCN-AE) proposed an autoencoder architecture for anomaly detection in time series which is based on temporal convolutional networks. \citeauthor{shen2020timeseries}\cite{shen2020timeseries}(THOC) proposed a temporal one-class classification model based on a dilated recurrent neural network with skip connections for time series anomaly detection. \citeauthor{li2021multivariate}\cite{li2021multivariate}(InterFusion) simultaneously modeled the inter-metric and temporal dependency for multivariate time series anomaly detection. \citeauthor{garg2021evaluation}\cite{garg2021evaluation}(UAE) found that the simple independent channel-wise Univariate Fully-Connected AutoEncoder (UAE), when used with dynamic Gaussian scoring function, outperforms all other complex deep learning algorithms. They believe that the choice of scoring functions often matters more than the choice of the underlying model. This observation suggests that recently proposed deep learning algorithms need to be further improved for effectively detecting anomalies in multivariate time series datasets.

\subsection{Transformer-based anomaly detection}
Transformer \cite{vaswani2017attention} has achieved great success in natural language processing, computer vision, and many other fields. Several studies used the basic and modified attention mechanisms for time series data. \citeauthor{song2018attend}\cite{song2018attend} leveraged self-attention on medical time series data analysis. \citeauthor{ma2019cdsa}\cite{ma2019cdsa} leveraged self-attention on estimating missing time series values. \citeauthor{wu2020deep}\cite{wu2020deep} used a Transformer-based model to forecast influenza-like illness. \citeauthor{Cohen2020TransformerBasedAS}\cite{Cohen2020TransformerBasedAS} proposed Transformer-based anomaly segmentation to detect anomalous images. \citeauthor{zerveas2021transformer}\cite{zerveas2021transformer} proposed a transformer-based framework for unsupervised representation learning of multivariate time series. \citeauthor{zerveas2021transformer}\cite{zaheer2020big} pointed out one of the core limitations of transformers: the quadratic dependency (mainly in terms of memory) on the sequence length due to their full attention mechanism. To remedy this, they proposed a sparse attention mechanism that reduced the quadratic dependency to linear. \citeauthor{chen2022net}\cite{chen2022net} introduced a specific Swin Transformer which integrated two attention mechanisms, Squeeze-Excitation Window Attention (SEWA) and Sparse Self-Attention within Windows (SSAW) to multi-task time series classification to handle global long-range dependencies.

For time series anomaly detection, \citeauthor{xu2021anomaly}\cite{xu2021anomaly} proposed the Anomaly Transformer to model temporal association and used a minimax strategy to amplify the normal-abnormal distinguishability of the Association Discrepancy to detect the anomalies. \citeauthor{tuli2022tranad}\cite{tuli2022tranad}(TranAD) presented a Transformer-based anomaly detection model and leveraged self-conditioning and adversarial training to amplify errors and gained training stability.

Transformer-based methods have not yet been widely adapted in multivariate time series, especially for anomaly detection, due to reasons such as the memory bottleneck, difficulty in encoding positional information, Transformer's focuses on pointwise representation and pairwise association, etc.

\subsection{Graph neural network-based anomaly detection}
In recent years, graph neural networks (GNNs) have emerged as successful approaches for modeling complex relationships between sensors. \citeauthor{zhao2020multivariate}\cite{zhao2020multivariate}(MTAD-GAT) introduced the graph attention (GAT) layers to learn the complex dependencies of multivariate time series. \citeauthor{deng2021graph}\cite{deng2021graph}(GDN) proposed graph structure learning by computing the similarity of embedding vectors between a sensor and all other sensors. \citeauthor{chen2021learning}\cite{chen2021learning}(GTA) proposed to learn a graph structure automatically using the Gumbel-Softmax sampling strategy and used influence propagation convolution to model the information flow between graph nodes. However, the graph structures learned by the models above are spatial topologies, which are static.

\section{Methodology}
In this section, we will introduce our model, EdgeConvFormer, for anomaly detection in multivariate time series. The key structure is an stack of multiple layers of EdgeConv module and Transformer module, facilitating the learning of the potential spatiotemporal topology and temporal association from multi-scale features.

\subsection{Problem definition}
Given the training historical data  \begin{math}X_{train}\in\end{math} \begin{math} \mathbb{R}^{T_{1}\times S} \end{math} of \begin{math}S\end{math} time series (sensors) with \begin{math}T_{1}\end{math} timestamps, and assuming that there is no anomaly in the training data, the task is to predict whether an anomaly occurred at each timestamp \begin{math}t\end{math} in the unseen test time-series \begin{math}X_{test}\in\end{math} \begin{math} \mathbb{R}^{T_{2}\times S} \end{math} with \begin{math}T_{2}\end{math} timestamps.  \begin{math}x_{t}\end{math} is used to denote the data point at a timestamp \begin{math}t\end{math} where \begin{math}x_{t}\in\end{math} \begin{math} \mathbb{R}^{S} \end{math}. \begin{math}X^{i}\end{math} is used to denote the time series of sensor \begin{math}i\end{math}. We convert \begin{math}X_{train}\end{math} and \begin{math}X_{test}\end{math} into overlapping windows of length \begin{math}l_{w}\end{math} with stride \begin{math}l_{s}\end{math}. For simplicity, the batch dimension is neglected throughout the introduction. We adopt a reconstruction-based strategy to predict the reconstruction of each input time-series window. The reconstruction error is denoted as \begin{math}Er_{t}^{i}\end{math} for sensor \begin{math}i\end{math}, at timestamp \begin{math}t\end{math}. We apply the dynamic Gaussian scoring function \cite{garg2021evaluation,ahmad2017unsupervised} based on \begin{math}Er_{t}^{i}\end{math} to calculate the sensor-wise anomaly score \begin{math}a_{t}^{i}\end{math} at timestamp \begin{math}t\end{math}, and then add all the sensor-wise anomaly scores, resulting in the final anomaly score \begin{math}A_{t}\end{math} at timestamp \begin{math}t\end{math}. Finally, we use different thresholding functions including best-\begin{math}F\end{math}-score, Top-k, and Tail-p to obtain binary labels \begin{math}y(t)\in\{0, 1\}\end{math} to indicate whether there is an anomaly at timestamp \begin{math}t\end{math}.

\begin{figure}[ht]
  \centering
  \begin{subfigure}[b]{0.9\textwidth}
    \includegraphics[width=\textwidth]{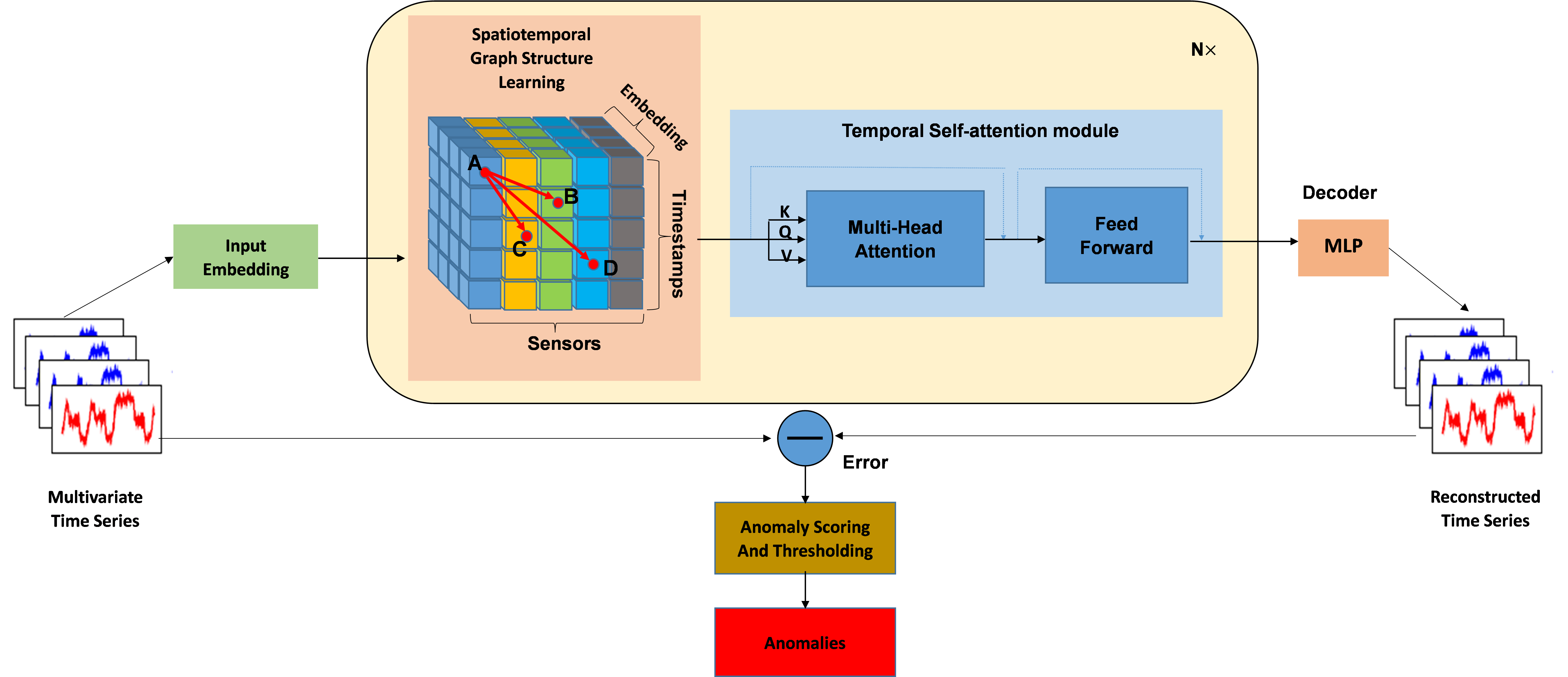}
    \caption{EdgeConvFormer Model Overview}
    \label{fig:subfig1}
  \end{subfigure}
  \begin{subfigure}[b]{0.9\textwidth}
    \includegraphics[width=\textwidth]{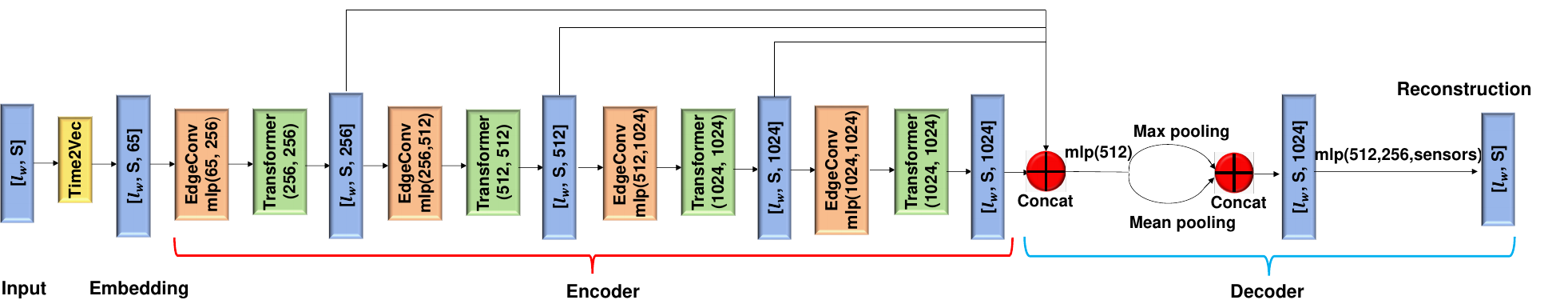}
    \caption{EdgeConvFormer Backbone}
    \label{fig:subfig2}
  \end{subfigure}  
  \caption{EdgeConvFormer Architecture. (1) Input Embedding: The Time2Vec module extracts periodic and complex patterns of time information for each channel (sensor), independently. (2) Encoder: EdgeConv module and Transformer module are integrated in a hierarchical, multi-scale manner and reinforce each other at each layer. EdgeConv provides more meaningful information to the Transformer by extracting the embeddings most relevant to the center point from the nearest neighbors in the two-dimensional space (sensors, timestamps). Meanwhile, focusing on the time dimension with the Transformer enhances EdgeConv's ability to capture information across long-term timestamps. (3) Decoder: An MLP layer is used to aggregate these multi-scale features extracted by the encoder and project to the sensor dimension.}
  \label{fig:architecture}
\end{figure}

\subsection{Network architecture}
As shown in Fig.~\ref{fig:architecture} (a), we propose an encoder-decoder reconstruction-based anomaly detection model, EdgeConvFormer, which encodes the input embeddings by using Time2Vec to capture both the periodic and aperiodic patterns in temporal information. It also makes full use of local space-time topology by constructing the local neighborhood graph using convolution-like operations on the edges of adjacent sensor pairs in the space-time two-dimensional space to update the embeddings. Transformer is used on this basis to discover long-term time dependencies. The most informative features are extracted to reconstruct the input time series, and the time points with high reconstruction errors are identified as anomalies. The network architecture of our work is shown in Fig.~\ref{fig:architecture} (b). Time2Vec is used to generate embedding vectors for each sensor's time series separately and these embedding vectors are spliced into a three-dimensional space-time vector. In the encoder, the embedding vectors are fed into four layers of EdgeConv and Transformer with embedding sizes of 256,512,1024,1024 to progressively search the neighborhoods around each embedding vector in the space-time domain. This neighborhood information is fused to introduce local inductive biases to Transformer's self-attention across long-range timesteps. The decoder aggregates all the information to obtain the global features and reconstructs the input time series by the fully connected layers.

\subsection{Time2Vec embedding}
Different sensors can have very different and intricate temporal patterns, so we try to represent each sensor in a multi-dimensional way to capture the characteristic `factors' underlying its behaviors. Time series decomposition is a standard method in time series analysis \cite{cohen1995time}. \citeauthor{mehran2019time2vec} \cite{mehran2019time2vec} proposed Time2vec, a model-agnostic vector representation for time, which can capture both the periodic and non-periodic patterns and is invariant to time rescaling. This approach is related to time decomposition techniques that encode a temporal signal into a set of frequencies, and its novelty is that it allows the frequencies to be learned instead of the common practice of using a fixed set of frequencies. Given the sensor-wise input time series data \begin{math}X^{i}\end{math}, the Time2Vec converts it to a ($m+1$)-dimensional embedding as follows:
  \begin{equation}
t2v\left(X^{i}\right)[j]=\left\{
\begin{aligned}
 \omega_{j}X^{i}+\varphi_{j},  \quad  & if\quad  j=0  \\
 \mathcal{F}\left(\omega_{j}X^{i}+\varphi_{j}\right),  \quad & if\quad  1\leq j \leq m.
\end{aligned}
\right.
\end{equation}
where \begin{math}\mathcal{F}\end{math} is a periodic activation function, \begin{math}\boldsymbol{\omega}\end{math} and \begin{math}\boldsymbol{\varphi}\end{math} are learnable parameters, which are the weight coefficients in the Time2vec embedding layer. In this study, $m$ is set to 64 as suggested by \cite{mehran2019time2vec}, and $\mathcal{F}$ is set to be the sine function. Note that the sine function with different parameters will capture the periodic patterns of various frequencies while the linear term (when $j=0$) extracts the non-periodic pattern of the time series.

The Time2Vec embedding is performed individually for each sensor and the per-sensor results are concatenated along the sensor dimension. Thus, the output shape of this module is \begin{math}[l_{w}, S, 64+1]\end{math} , where \begin{math}l_{w}\end{math} denotes the length of the subsequence of each sliding window, \begin{math}S\end{math} denotes the number of sensors, and \begin{math}64+1\end{math} is the dimension of the embedding, which represents 64 sine functions and 1 linear term.

\subsection{Encoder}
As shown in Fig.~\ref{fig:architecture} (b), the encoder consists of four layers of EdgeConv+Transformer with different output dimensions connected end-to-end. The EdgeConv used to recover the topological information in point clouds \cite{wang2019dynamic}, can dynamically construct a graph structure on each layer of the network. We deploy an EdgeConv module and a Transformer module, and the Transformer module is used to capture the information across long-distance timestamps in each layer and stack four such layers. The representation is carried out by continuously stacking the representation results of each layer of EdgeConv-Transformer, i.e., the output of the previous EdgeConv-Transformer layer is the input of the next EdgeConv-Transformer layer, except that the first layer's input comes from the output of the Time2Vec module.

\textbf{\textit{EdgeConv module:}} At each layer, the input of EdgeConv is reshaped as  \begin{math}[l_{w}\times S, embedding]\end{math}. Here \begin{math}l_{w}\times S \end{math} means the EdgeConv performs on the points of the two-dimensional (space-time) space. The \begin{math}embedding\end{math} is the embedding vector of each point in this two-dimensional space. Each embedding layer has different dimensions: The first layer is 65 because it comes from the output of Time2Vec. In the remaining layers, the embeddings are from the output of the Transformer of the previous layer. The network architecture of our EdgeConvFormer consists of four layers, and the dimensions of the input embedding of EdgeConv in each layer are 65, 256, 512, and 1024 respectively. We can express the embedding vector (dubbed as edge features in a graph) of each point on the \begin{math}l_{w}\times S \end{math} two-dimensional space of layer \begin{math}l\end{math} as:
\begin{equation}
\begin{aligned}
h_{i}^{(l)} \in \mathbb{R}^{d_{l}}, \quad for \quad i\in\{1,2,...l_{w} \times S\} \\
 where \quad d_{l}\in\{65,256,512,1024\}.
\end{aligned}
\end{equation}

The EdgConv of each layer firstly constructs the \begin{math}k\end{math}-nearest neighbor graph for all points in the two-dimensional space (\begin{math}l_{w}\times S \end{math}) by calculating the Euclidean distance of the edge features. The graph includes a self-loop, meaning each node also points to itself. The EdgConv operates on each point respectively and looks at it as a center point, and aggregates its edge feature \begin{math} h_{i}^{(l)}\end{math} with neighbor's edge \begin{math} h_{j}^{(l)}\end{math} to obtain a new representation \begin{math} h_{i}^{(l+1)}\end{math} of this point. The process can be described as follows:
\begin{equation}
h_{i}^{\left(l+1\right)}=\max_{j\in N\left(i\right)}\left(ReLU\left(\Theta\cdot\left(h^{\left(l\right)}_{j}-h_{i}^{\left(l\right)}\right)+\Phi\cdot h_{i}^{(l)}\right)\right),
\end{equation}
where \begin{math} N(i)\end{math} is the set of the \begin{math}k\end{math}-nearest neighbours of \begin{math}i\end{math}; and \begin{math}\Theta\end{math} and \begin{math}\Phi\end{math} are linear layers. The obtained edge features combine the previous layer's global information \begin{math}h_{i}^{(l)}\end{math} and the local neighborhood information  \begin{math}h^{(l)}_{j}-h_{i}^{(l)}\end{math}. Max pooling is an aggregate operation to extract the features of an edge with the largest correlation regarding the center point. The dimensions of \begin{math}h_{i}^{(l+1)}\end{math} of the four EdgeConv layers are: 256, 512, 1024, and 1024 respectively.

\begin{figure}[ht]
\centering
\includegraphics[width=0.8\textwidth]{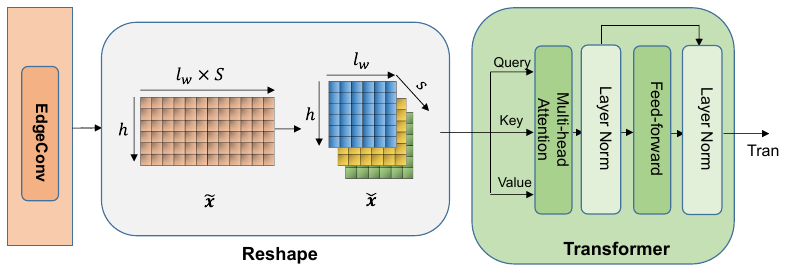}

\caption{Self-attention on time dimension}
\label{fig:attention}
\end{figure}

\textbf{\textit{Transformer  module:}} As shown in Fig.~\ref{fig:architecture}, we add a Transformer module after the EdgeConv module in each layer to enhance the representation power of features by capturing the information across long-range timestamps. We use the encoder part of the Vanilla Transformer but remove the position embedding as the sequence order information has already been extracted in the previous Time2Vec module. As shown in Fig. ~\ref{fig:attention}, at each layer, the output of EdgeConv \begin{math}\widetilde{x}\end{math} is a matrix whose rows correspond to points \begin{math}i\in\{1,2,...l_{w} \times S\}\end{math} and columns correspond to edge features \begin{math}h_{i}^{(l+1)}\end{math}. We firstly reshape \begin{math}\widetilde{x}\end{math} into a 3-dimensional vector \begin{math}\check{x}:  [S, l_{w}, h]\end{math}. Next, multiple sensors are processed in parallel by tensor broadcast, which means for each sensor, self-attention is used to operate on the matrix \begin{math}[l_{w}, h]\end{math}. The relationship between the embedding \begin{math}h \end{math} of each timestamp and the embedding \begin{math}h\end{math} of the other timestamps in the entire time window \begin{math}l_{w}\end{math} are established. An encoded representation for each embedding is hence produced that incorporates the attention scores for each embedding in the input time window \begin{math}l_{w}\end{math}.

In each layer, the Transformer module performs the following operations. Firstly, the query (Q), key (K), and value (V) matrices are calculated by multiplying \begin{math}\check{x}\end{math} with the weight matrices \begin{math}w^{Q}, w^{K}, w^{V}\end{math} created during training
(the three matrices are initialized randomly and learnable).
\begin{equation}
Q = \check{x} \times w^{Q} , \quad
K = \check{x} \times w^{K} , \quad
V = \check{x} \times w^{V} .
\end{equation}

Next Multi-Head Scaled Dot-Product Attention is adopted. Eq. \ref{equ:attention} is from \cite{vaswani2017attention}:

\begin{equation}
\begin{aligned}
\textit{Attention}\left(Q,K,V\right) = \textit{softmax}\left(\frac{QK^{T}}{\sqrt{d_{k}}}\right)V , \\
\textit{MultiHeadAtt}\left(Q,K,V\right) = \textit{Concate}\left(H_{1},...,H_{h}\right), \\
 \textit{where} \quad  H_{i} = \textit{Attention}\left(Q_{i},K_{i},V_{i}\right).
\end{aligned}
\label{equ:attention}
\end{equation}

A residual connection is used here, followed by layer normalization. The formula is as follows:
\begin{equation}
\begin{aligned}
\textit{Tr} &= \textit{LayerNorm}\left(\check{x} + \textit{MultiHeadAtt}\left(Q,K,V\right)\right),  \\
\textit{Tran} &= \textit{LayerNorm}\left(\textit{Tr}+\textit{FeedForward(Tr)}\right).
\end{aligned}
\end{equation}

A Feedforward layer is used here to output different feature dimensions. The output shape of the Transformer module is \begin{math} [S, l_{w}, embedding]\end{math}.

Due to the parallel computation of Self-attention, the above operations can be performed over multiple batches of the time-series windows and multiple sensors in parallel and speed up the training process \cite{tuli2022tranad}.

\subsection{Decoder}
The decoder part is shown in Fig.~\ref{fig:architecture} (b). The outputs of the four-layer EdgeConv-Transformer module with multi-scale features (256, 512, 1024, 1024) are shortcut-connected and spliced. An MLP layer is used to aggregate these features and reduce the dimension of features to 512. We use both the max pooling and average pooling on sensor dimension \begin{math}S \end{math} and concatenate their results on the feature dimension. The feature dimension is changed to 1024 to obtain global features. Finally, three fully connected layers are used with dropout, LeakyReLU, and layer normalization. The last layer of the decoder is a linear projection whose number of output channels equals the number of sensors, resulting in the final reconstructed time-series window \begin{math}\hat{x}\end{math} with shape \begin{math}[l_{w}, S]\end{math}. For the loss function, we use the mean squared error (MSE) between the reconstructed and original signals in the time-series window, that is,
\begin{equation}
L_{\textit{MSE}} = \frac{1}{l_{w}}\sum\limits_{t=0}^{l_{w} - 1}\lVert\hat{x}_{t} - x_{t}\rVert_{2}^{2}.  \\
\end{equation}

A reconstruction error at time \begin{math}t\end{math} for sensor \begin{math}i\end{math} is obtained by calculating the absolute difference between the reconstructed value and the original ones and used as the basis to judge the degree of abnormality:
\begin{equation}
\textit{Er}_{t}^{i} = \lVert\hat{x}_{t}^{i} - x_{t}^{i}\rVert_{1}.
\end{equation}

\section{Anomaly detection and evaluation}
While the reconstruction error reveals the anomaly degree to some extent, sometimes it does not distinguish abnormal from normal data clearly. A so-called scoring function is needed to enlarge the anomaly score disparity for better detection \cite{huang2021enhancing}. For a hard detection of an anomaly, a thresholding method is applied to the anomaly score for a binary decision. The threshold method can help detect anomalies more accurately by reducing false positives and false negatives \cite{halbe2017abnormal}). Furthermore, evaluation metrics are also needed to measure the performance of different algorithms on a given data set.

\subsection{Anomaly scoring function}
\textbf{\textit{Gauss\_D}}: We adopt the dynamic Gaussian scoring function which adapts to variations in the test set, as it outperforms static scoring functions \cite{ahmad2017unsupervised}. We fit a Gaussian distribution with sample mean \begin{math}\mu_{t}^{i} \end{math} and variance \begin{math}\sigma_{t}^{i}\end{math} (where \begin{math}i\end{math} is the index for sensors, and \begin{math}t\end{math} is the index for timestamps), continuously updated by error values from the previous sliding window. Using the mean and variance computed from the last sliding window of the training set as the mean \begin{math}\mu_{t  }^{i} \end{math} and variance \begin{math}\sigma_{t}^{i}\end{math} of the initial part of the test set, this is the so-called rolling normal distribution \cite{garg2021evaluation}. This process can be expressed as follows:
\begin{equation}
\begin{aligned}
\mu_{t}^{i} &= \frac{1}{l_{w}}\sum\limits_{j=0}^{l_{w} - 1}Er_{t - j}^{i}  \\
\left(\sigma_{t}^{i}\right)^{2} &= \frac{1}{l_{w} - 1}\sum\limits_{j=0}^{l_{w} - 1}\left(Er_{t - j}^{i} - \mu_{t}^{i}\right)^{2}.
\end{aligned}
\end{equation}

Next, we calculate the sensor-wise anomaly score \begin{math}a_{t}^{i}\end{math} based on the fitted distribution using the cumulative distribution function (\textit{cdf}): \begin{math}- \log(1 - cdf )\end{math}, which increases monotonically with reconstruction error \cite{garg2021evaluation}. It can be expressed as Eq.~\ref{eq:score1}, in which \begin{math}\Phi\end{math} denotes the \textit{cdf} of \begin{math}N (0, 1)\end{math}:
\begin{equation}
a_{t}^{i} = - \log\left( 1 - \Phi\left( \frac{Er_{t}^{i}\ - \mu_{t}^{i}}{\sigma_{t}^{i}}\right)\right).
\label{eq:score1}
\end{equation}

As shown in Eq.~\ref{eq:score2}, adding all these sensor-wise anomaly scores results in the final anomaly score \begin{math}A_{t}\end{math} at timestamp \begin{math}t\end{math}:
\begin{equation}
A_{t} = \sum\limits_{i=1}^{N}a_{t}^{i}.
\label{eq:score2}
\end{equation}

\textbf{\textit{Gauss\_D\_K}}: The dynamic Gaussian scoring function with Gaussian kernel, applies Gaussian kernel convolution based on \begin{math}Gauss\_D\end{math} scoring \cite{garg2021evaluation}. Since the total anomaly score is the sum of the anomaly scores of all sensors at each time point, and sometimes multiple sensors responding to the same anomaly event will appear at slightly different time points, this maybe produces misleading spikes. The gaussian kernel can smooth the anomaly scores across these sensors and remove noise such that the total anomaly score is more accurate. The way to do this is to use a two-dimensional Gaussian kernel of a certain size to convolve with the channel-wise anomaly scores from \begin{math}Gauss\_D\end{math}. It can be expressed as Eq.~\ref{eq:score3}:
\begin{equation}
\begin{aligned}
G\left(\mu;\sigma_{k}\right) &=e^{-\frac{1}{2}\left(\frac{\mu}{\sigma_{k}}\right)^{2}}  \\
a_{t;\textit{Gauss}\_D\_K}^{i} &= G*a_{t;\textit{Gauss}\_D}^{i}.
\end{aligned}
\label{eq:score3}
\end{equation}
where \begin{math}G\end{math} is a Gaussian filter with kernel sigma \begin{math}\sigma_{k}\end{math} ( a parameter that controls the width of the Gaussian function and needs to be tuned according to different data sets), * denotes the convolution operator and \begin{math}a_{t;Gauss\_D}^{i}\end{math} refers to the channel-wise anomaly scores from \begin{math}Gauss\_D\end{math}.

\subsection{Threshold method}
Many recent anomaly detection algorithms use Best-\begin{math}F_{1}\end{math}-score threshold \cite{xu2021anomaly}. \citeauthor{garg2021evaluation}\cite{garg2021evaluation} used three thresholding functions: Best-\begin{math}F\end{math}-score, Top-\begin{math}k\end{math}, and Tail-\begin{math}p\end{math} to evaluate algorithm performances. For a fair comparison with their algorithms, we also use these three thresholds as defined below:

\textbf{\textit{Best-\begin{math}F\end{math}-score}}
 The $Best$-$F$-$score$ threshold is used to search for anomaly threshold under the best \begin{math}F\end{math} score: the maximum value of the used evaluation metric ($F_{1}$,  $Fpa_{1}$ or $Fc_{1}$, which will be introduced in the next section).

\textbf{\textit{Top-k}}
The $Top$-$k$ threshold is to select $k$ time-points with the highest anomaly scores and label them as anomalous, where $k$ is the actual number of anomalies and varies in different test sets.

\textbf{\textit{Tail-p}}
Since our scoring function is a sum of negative log probabilities of \begin{math}N\end{math} sensors, we use the $Tail$-$p$ threshold to label time-points with scores \begin{math}A_{t}> - N\log_{10}(\epsilon)\end{math} as anomalous, where \begin{math}N\end{math} is the number of sensors, and \begin{math}\epsilon\end{math} is a small tail probability, \begin{math}\epsilon \in \{10^{-1},10^{-2},10^{-3},10^{-4},10^{-5}\}\end{math}. The threshold is set by using these five values for \begin{math}\epsilon\end{math} separately and selecting the value with the highest performance.

Among the three methods, $Tail$-$p$ is more consistent with the streaming scenario and more applicable in practical applications. Best-\begin{math}F\end{math}-score  needs to mark the label of the complete test set in advance, while $Top$-$k$ requires the abnormal score of the complete test set. In practice, we have found that after tuning an appropriate \begin{math}\epsilon\end{math} for $Tail$-$p$ in an application, this \begin{math}\epsilon\end{math} can be applied for all subsequent data streams in the same application.

\subsection{Evaluation metrics}
\textbf{\textit{\pmb{$F_{1}$} score:}} $F_{1}$ score is the harmonic mean of point-wise precision and recall. It is a traditional metric to detect point anomalies, and it accounts for both false positives and false negatives.

\textbf{\textit{\pmb{$Fpa_{1}$} score:}} Recent state-of-the-art methods in time series anomaly detection \cite{xu2018unsupervised,zhou2019beatgan,su2019robust,li2021multivariate,shen2020timeseries,xu2021anomaly} used point-adjusted F$_{1}$ (\begin{math}Fpa_{1}\end{math}) score to evaluate the detection performance. All instances in an anomalous segment are considered true positives if a single anomaly is detected in the entire segment. The idea is consistent with how abnormalities are handled in real-world applications, i.e., it is acceptable to trigger an alarm at any point in a continuous abnormal segment, as long as the delay is not too long. However, it leads to an inflated score as the false detection is only penalized once, while the correct detection is generously rewarded \cite{doshi2022tisat}.

\textbf{\textit{\pmb{$Fc_{1}$} score:}} \citeauthor{garg2021evaluation}\cite{garg2021evaluation} proposed a new evaluation metric - the Composite \begin{math}F_{1}\end{math} score (\begin{math}Fc_{1}\end{math}), which is a combination of \begin{math}F_{1}\end{math} and \begin{math}Fpa_{1}\end{math}.  They think \begin{math}F_{1}\end{math} is a point-wise score, which may be too pessimistic since triggering an alarm for each abnormal point is not practical in real applications. \begin{math}Fpa_{1}\end{math}, on the other hand, is the segment-wise score, which treats all anomalous points in a segment as detected and  counts all such anomalous points as true positives (TP) as long as one anomalous point is detected, but only counts one false negative (FN) if none of the anomalous points are detected in that segment, which is too optimistic. \begin{math}Fc_{1}\end{math} is a compromise between \begin{math}F_{1}\end{math} and \begin{math}Fpa_{1}\end{math} by calculating the harmonic mean of the point-wise precision, \begin{math}P_{t}\end{math}, and the segment-wise recall, \begin{math}R_{s}\end{math}, as follows:
\begin{equation}
\begin{aligned}
Fc_{1} &= \frac{2\times P_{t}\times R_{s}}{P_{t} + R_{s}} \\
where \quad P_{t} &= \frac{TP_{t}}{TP_{t}+FP_{t}}, R_{s} = \frac{TP_{s}}{TP_{s}+FN_{s}} .
\end{aligned}
\end{equation}

\begin{math}TP_{t}\end{math} and \begin{math}FP_{t}\end{math} are the point-wise true positive and false positive respectively, while \begin{math}TP_{s}\end{math} and \begin{math}FN_{s}\end{math} are the segment-wise true positive and false negative respectively. \begin{math}Fc_{1}\end{math} can more rationally judge how far a detector is from having perfect detection ability: detects at least one point in each abnormal segment (\begin{math}R_{s} = 1\end{math}) and has no point-wise false positive \begin{math}FP_{t}\end{math} (\begin{math}P_{t} = 1\end{math}).

\textbf{\textit{\pmb{$F_{1}$}\pmb{-{$P_{T}$$R_{T}$}} score:}} Many real-world anomalies are range-based, i.e., they occur over a period of time, and they constitute a subset of both contextual and collective anomalies \cite{chandola2009anomaly}. $F_{1}$-$P_{T}$$R_{T}$ is the harmonic mean of respective ranged-based precision ($P_{T}$) and range-based Recall ($R_{T}$) values, which are more recent and more reasonable evaluation metrics specifically developed for range-based time series anomalies \cite{tatbul2018precision}. $P_{T}$ and $R_{T}$ of the whole time series are the average of range-based precision and range-based recall of all real abnormal segments, respectively. For each real abnormal segment, the calculation of the range-based precision and range-based recall takes into account four aspects: 1) Existence: at least a single anomaly is detected in the entire segment, same as \begin{math}Fpa_{1}\end{math}; 2) Size: the wider a range of an anomaly that an anomaly detection algorithm can detect, the better its performance; 3) Position: the earlier an anomaly is first flagged in this  abnormal segment, the better the algorithm; 4) Cardinality: only one fragment should be detected in this abnormal segment instead of multiple fragments, as duplicate alerts for a single anomaly event are redundant and confusing. The four aspects obove are represented by four tunable parameters $\alpha$ (Existence), $\omega$ (Size), $\delta$ (Position), and $\gamma$ (Cardinality) respectively. Please refer to the original paper for details. \citeauthor{jacob2020exathlon}\cite{jacob2020exathlon} set the above four aspects as four-level criteria for functional evaluation of anomaly detection algorithm: AD1 (Anomaly Existence), AD2 (Range Detection), AD3 (Early Detection), AD4 (Exactly-Once Detection), and a higher AD level includes the requirements of all preceding levels, and $F_{1}$-$P_{T}$$R_{T}$ score (AD1) $\geq$ $F_{1}$-$P_{T}$$R_{T}$ score (AD2) $\geq$ $F_{1}$-$P_{T}$$R_{T}$ score (AD3) $\geq$ $F_{1}$-$P_{T}$$R_{T}$ score (AD4).

\textbf{\textit{\pmb{$AU$-$ROC$}, \pmb{$AU$-$PRC$}:}} $AU$-$ROC$ (area under the receiver operator characteristic curve), is the metric that indicates the trade-off in the true-positive rate and false-positive rate for different thresholds. The closer the $AU$-$ROC$ is to 1, the better the performance. $AU$-$PRC$ (area under the precision-recall curve) focuses on the performance of a detector on the positives only. $AU$-$PRC$ is a critical metric used in scenarios where correctly classifying the positives is important \cite{fernandez2018learning}.

\citeauthor{garg2021evaluation}\cite{garg2021evaluation} claimed that the choice of scoring functions often matters more than the choice of the underlying model. Their finding is that the simple Univariate Fully-connected AutoEncoder (UAE), when used with the dynamic Gaussian scoring function, outperforms all other complex algorithms for anomaly detection. To prove that the choice of the underlying model is equally important, we compare EdgeConvFormer with the UAE model and other top-ranked models in Garg's paper, such as LSTM-VAE, MSCRED, TCN-AE, OmniAnomaly, and BeatGan. We employ dynamic Gaussian scoring and three threshold methods: best-F-score, top-k, and tail-p. Since there is currently no commonly accepted evaluation metric, we use $F_{1}$ score, $Fpa_{1}$ score, and $Fc_{1}$ score, and another two overall metrics $AU$-$ROC$ and $AU$-$PRC$ to conduct a comprehensive evaluation of the algorithms.

There is also some recently published work with no source code available \cite{shen2020timeseries, xu2021anomaly}. \begin{math}Fpa_{1}\end{math} is used as their evaluation metric. The results reported by the authors are included in our comparison. To unify the evaluation metrics, we use precision, recall, and \begin{math}Fpa_{1}\end{math} score to evaluate the detection performance of these works.

Exathlon \cite{jacob2020exathlon} is a new benchmark for range-based anomalies detection over high-dimensional time series data (i.e., multivariate with 2,283 dimensions), constructed based on real data traces from repeated executions of large-scale stream processing jobs on an Apache Spark cluster. We adopt range-based $F_{1}$-$P_{T}$$R_{T}$ with four-level to demonstrate that our proposed method is equally effective in this dataset.

\begin{table*}[hbt]
\centering
\setlength\tabcolsep{2pt}
\caption{Five commonly-used datasets. Averaging is done over entities.}
\begin{tabular}{ccccccc}
   \toprule
   Benchmarks & Domain & Entities & Dimensions & Average train length  & Average test length & Average anomalies \% \\
   \midrule
   SMD & Server  & 28 & 38 & 25,300 & 25,301 & 4.21\% \\
   MSL & Spacecraft & 27 & 55 & 2,160 & 2,731 & 12.02\% \\
   SMAP & Spacecraft & 55 & 25 & 2,556 & 8,071 &12.79\% \\
   SWaT & Water Treatment & 1 & 51 & 473,400 & 414,569 & 12.14\% \\
   PSM & Server & 1 & 25 & 132,481 & 87,842 & 27.76\% \\
   \bottomrule
\end{tabular}
\label{Tab:dataset}
\end{table*}

\begin{figure}[ht]
  \centering
  \begin{subfigure}[b]{0.45\textwidth}
    \includegraphics[width=\textwidth]{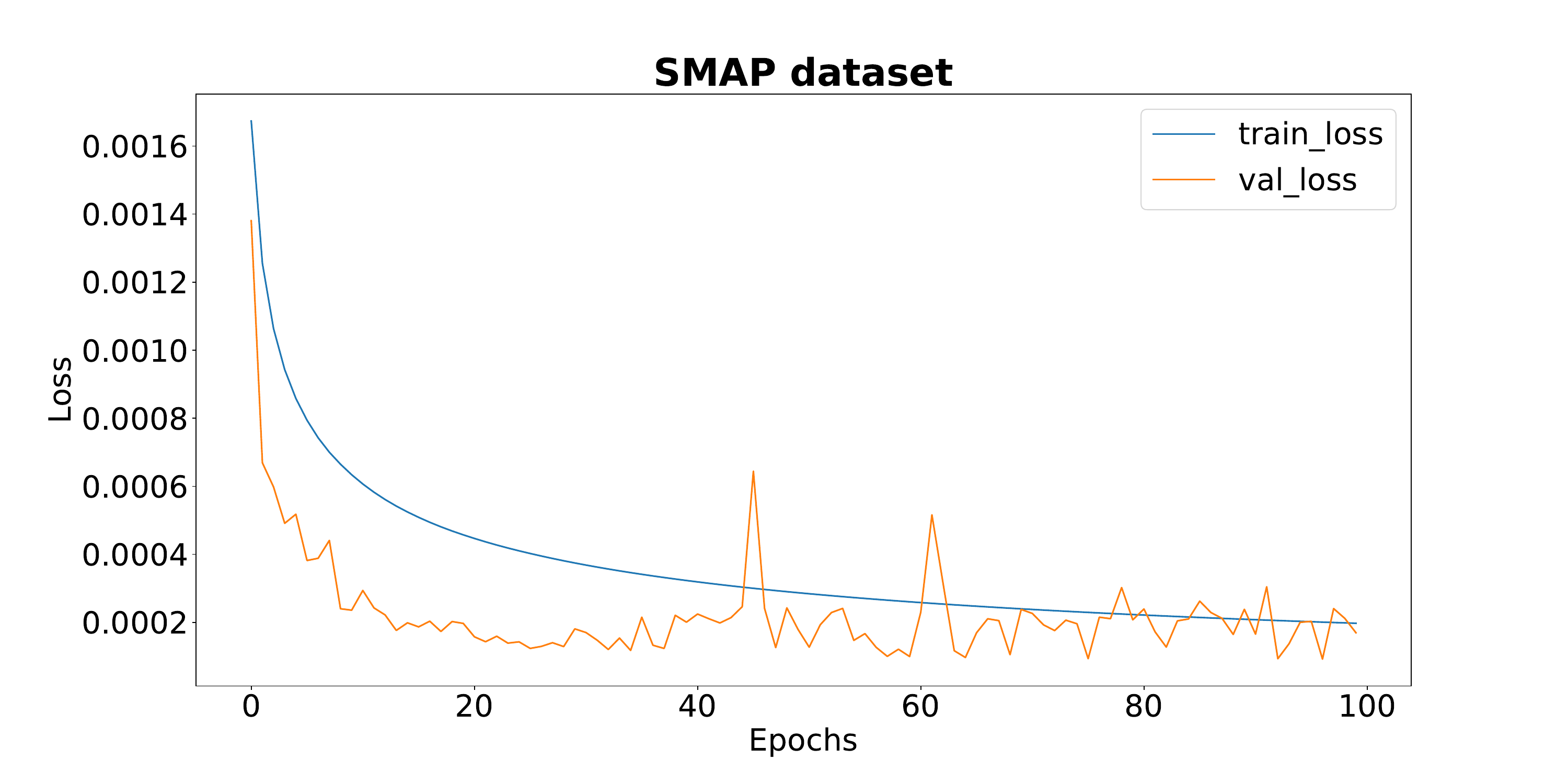}
    \label{fig:subfig4}
  \end{subfigure}
  \begin{subfigure}[b]{0.45\textwidth}
    \includegraphics[width=\textwidth]{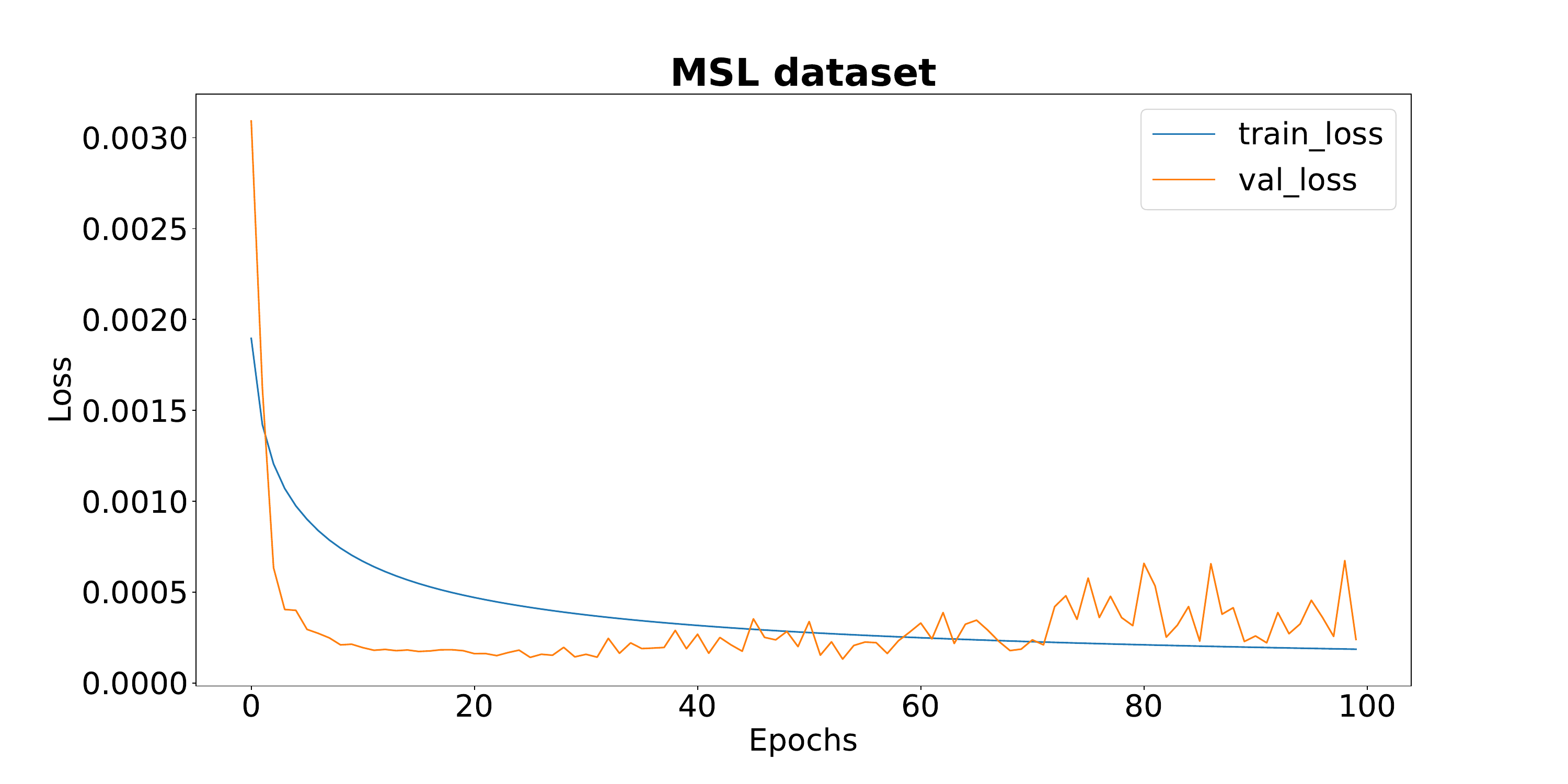}
    \label{fig:subfig5}
  \end{subfigure}  
  \begin{subfigure}[b]{0.45\textwidth}
    \includegraphics[width=\textwidth]{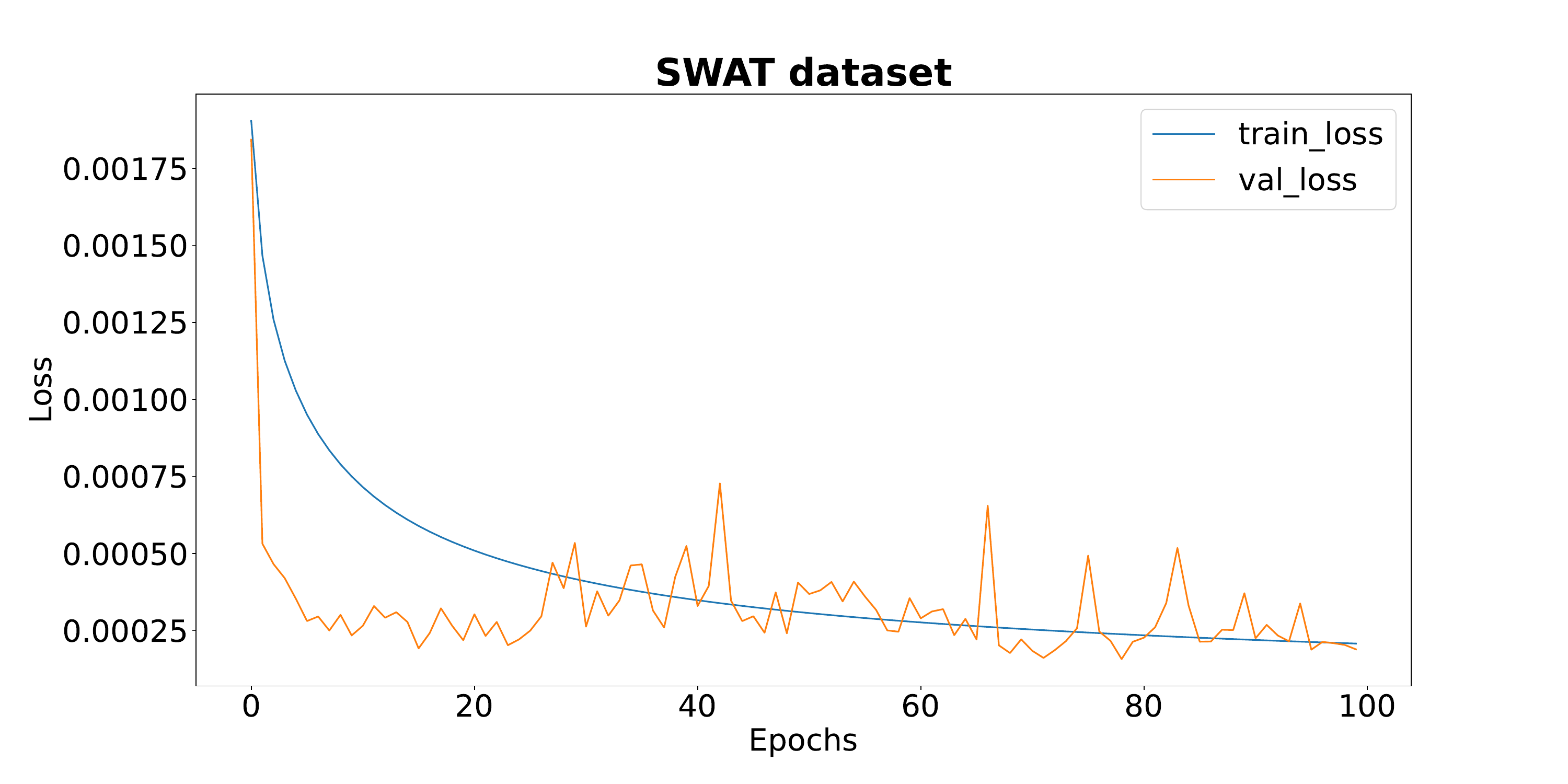}
    \label{fig:subfig6}
  \end{subfigure}  
  \begin{subfigure}[b]{0.45\textwidth}
    \includegraphics[width=\textwidth]{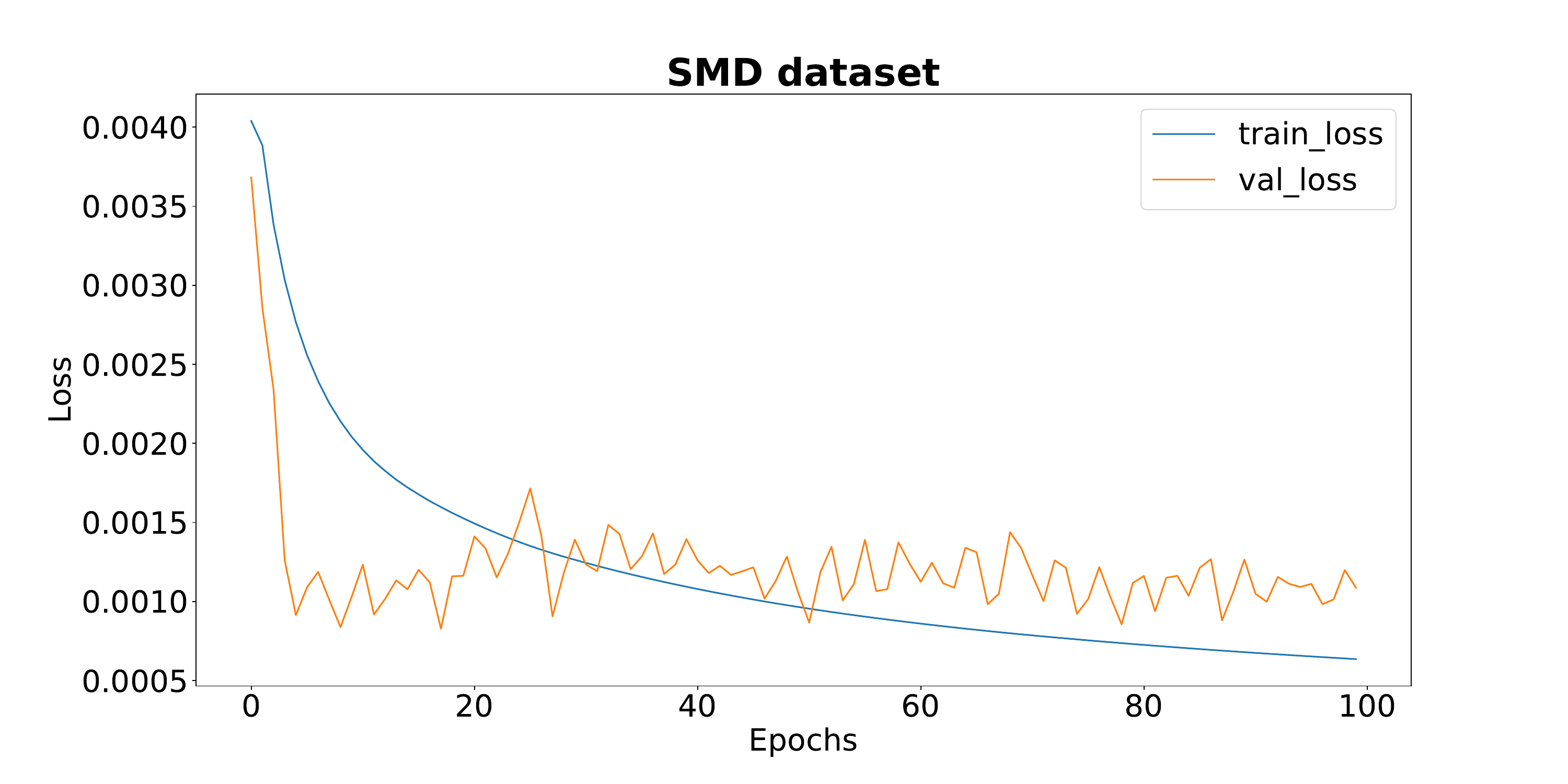}
    \label{fig:subfig7}
  \end{subfigure}  
  \begin{subfigure}[b]{0.45\textwidth}
    \includegraphics[width=\textwidth]{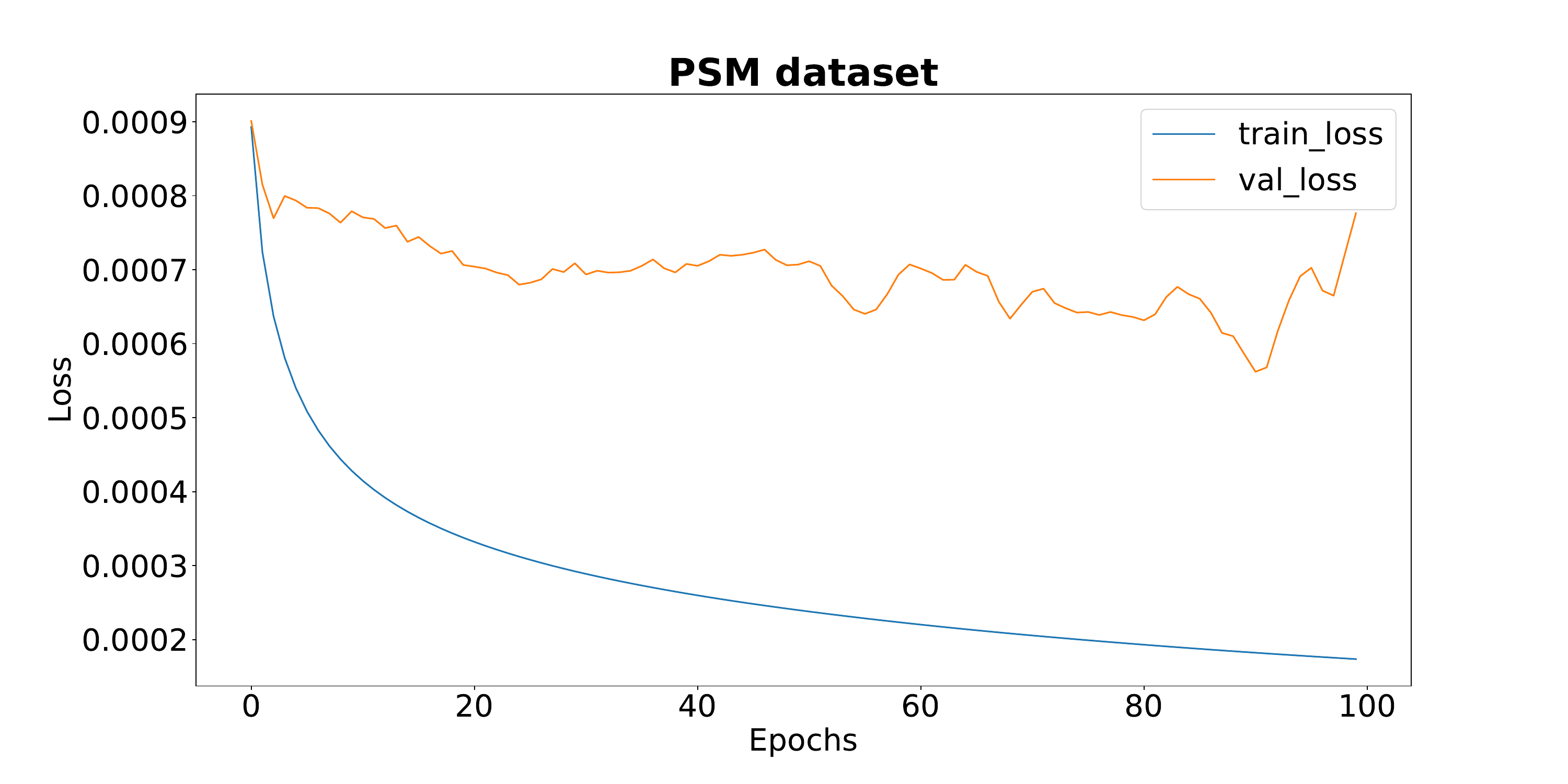}
    \label{fig:subfig8}
  \end{subfigure}  
  \caption{Train/Val loss of each dataset trained for 100 epochs.}
  \label{fig:loss}
\end{figure}

\section{Experiments}
\textbf{\textit{Five commonly-used datasets}}
We use 5 popular multivariate time series datasets from the real world (Table~\ref{Tab:dataset}) to evaluate the performance of our proposed EdgeConvFormer and to demonstrate its effectiveness in identifying unusual or unexpected patterns in the data. By applying our model to these datasets and comparing its performance to existing anomaly detection methods, we aim to show its potential for practical use in various domains and under different anomaly scenarios. (1) \textbf{SMD}  (Server Machine Dataset) \cite{su2019robust} is a 5-week-long dataset with 38 dimensions collected from a large Internet company. (2) \textbf{MSL} (Mars Science Laboratory rover) \cite{hundman2018detecting} consists of 55-dimensional telemetry data from NASA. (3) \textbf{SMAP} (Soil Moisture Active Passive satellite) \cite{hundman2018detecting} consists of 25-dimensional telemetry data from NASA. (4) \textbf{SWAT}  (Secure Water Treatment)\cite{goh2016dataset} is obtained from 51 sensors of a six-stage Secure Water Treatment testbed under 11 days of continuous operation, including 7 days of normal operation and 4 days of abnormal operation. (5) \textbf{PSM} (Pooled Server Metrics)\cite{abdulaal2021practical} is collected internally from multiple application server nodes at eBay with 26 dimensions. MSL, SMAP, and SMD are multi-entity datasets while the other two are single-entity datasets, where ``entity" means a different physical unit of the same type and same dimensionality. Similar to \cite{garg2021evaluation}, we train a separate model for each entity in the multi-entity datasets. The final evaluation is on the overall performance of detecting anomalies in all entities. The training sets are anomaly-free and test sets are anomaly labeled. The ratio of the training set and validation set is 8:2. Furthermore, to maintain the authenticity of the dataset and reduce the dependency on domain knowledge, we remove the special treatment of MSL, SMAP, and SWAT datasets in \cite{garg2021evaluation}, such as using only the reconstruction error of the first sensor in the MSL and SMAP datasets for anomaly detection, and reducing the longest abnormal segment to the average length of the abnormal segment in the SWAT dataset.

\textbf{\textit{Exathlon dataset}}
\citeauthor{wu2021current}\cite{wu2021current} recently pointed out that it is hard and unfair to compare anomaly detection methods on trivial benchmark datasets due to different definitions of anomaly and evaluation criteria. Therefore, we also tested our proposed model on the new much larger benchmark dataset, Exathlon, a high-dimensional and large-size dataset, utilizing the ranged-based Precision\&Recall\&F1 evaluation metrics to further evaluate our proposed method.

Exathlon is based on real data traces collected from around 100 repeated executions of 10 different Spark streaming applications on a 4-node cluster over 2.5 months. Exathlon provides a total of 93 traces, of which the trace is defined as a record of the Spark stream application running. These traces were divided into 59 undisturbed traces, which simply recorded the normal execution of the Spark streaming application, and 34 disturbed traces,  which recorded applications disturbed by events injected during known and labeled time intervals \cite{jacob2020exathlon}. As shown in TABLE~\ref{Tab:ExathlonTypes}, the dataset contains 5 different types of disturbed traces.

\begin{table}[]

\caption{Exathlon dataset \cite{jacob2020exathlon}.}

\begin{tabular}{c|l|c|c|c|l}
\hline
\textbf{\begin{tabular}[c]{@{}c@{}}Trace \\ Type\end{tabular}} & \multicolumn{1}{c|}{\textbf{\begin{tabular}[c]{@{}c@{}}Anomaly\\ Type\end{tabular}}} & \textbf{\begin{tabular}[c]{@{}c@{}}\# of\\ Traces\end{tabular}} & \textbf{\begin{tabular}[c]{@{}c@{}}Anomaly\\ Instances\end{tabular}} & \textbf{\begin{tabular}[c]{@{}c@{}}Anomaly length\\ min,avg,max\end{tabular}} & \multicolumn{1}{c}{\textbf{\begin{tabular}[c]{@{}c@{}}Data \\ Items\end{tabular}}} \\ \hline
Undisturbed                                                    & \multicolumn{1}{c|}{N/A}                                                             & 59                                                              & N/A                                                                  & N/A                                                                           & 1.4M                                                                               \\ \hline
Disturbed                                                      & T1:Bursty Input                                                                      & 6                                                               & 29                                                                   & 15m,22m,33m                                                                   & 360K                                                                               \\ \hline
Disturbed                                                      & T2: Bursty Input Until Crash                                                         & 7                                                               & 7                                                                    & 8m, 35m, 1.5h                                                                 & \multicolumn{1}{c}{31K}                                                            \\ \hline
Disturbed                                                      & T3: Stalled Input                                                                    & 4                                                               & 16                                                                   & 14m,16m,16m                                                                   & 187K                                                                               \\ \hline
Disturbed                                                      & T4: CPU Contention                                                                   & 6                                                               & 26                                                                   & 8m,15m,27m                                                                    & 181K                                                                               \\ \hline
Disturbed                                                      & T5: Process Failure                                                                  & 11                                                              & 19                                                                   & 1m,23m,2.8h                                                                   & 128K                                                                               \\ \hline
\end{tabular}

\label{Tab:ExathlonTypes}
\end{table}

\textbf{\textit{Implementation details}}
We use overlapping windows of length \begin{math}l_{w}\end{math} with stride \begin{math}l_{s}\end{math} to convert the dataset into a set of sub-sequences. The window size is set as \begin{math}l_{w} = 100\end{math} for all datasets. Because the training length of MSL and SMAP datasets is relatively short, we set the stride \begin{math}l_{s} = 1\end{math}, and for other larger datasets, we set the stride \begin{math}l_{s} = 10\end{math} to speed up training. All the parameters are tuned on the validation set and the minimum validation reconstruction error criterion is used.  ADAM optimizer is used, and the learning rate and the neighbor size\begin{math}k\end{math} are tuned for each dataset (see TABLE~\ref{Tab:para}).

As shown in Fig.~\ref{fig:loss}, our model has the characteristics of fast convergence on each dataset. To save computational cost, we only train for 3 epochs for all datasets except for PSM (epochs=20) as they have already achieved high detection accuracy. Four layers of EdgeConvFormer are implemented and \href{https://docs.dgl.ai/generated/dgl.nn.pytorch.conv.EdgeConv.html}{dgl} (Deep Graph Library) is used to apply EdgeConv to the spatial-temporal space and embedding space. The experiments on SMD, SMAP, and PSM are implemented on 2 NVIDIA TITAN RTX 24GB GPUs using multiple GPU training - \href{https://pytorch.org/tutorials/intermediate/FSDP_tutorial.html}{FSDP} (Fully Sharded Data Parallel).  The experiments on MSL, SWAT, and Exathlon are conducted on 4 NVIDIA TITAN RTX 24GB GPUs due to their larger dimensions.

As shown in TABLE~\ref{Tab:baseline}, all the baseline models except OmniAnomaly are trained for 100 epochs. The parameters of these models are consistent with those in \cite{garg2021evaluation}.

\begin{figure}[]
\centering
\includegraphics[width=0.5\textwidth]{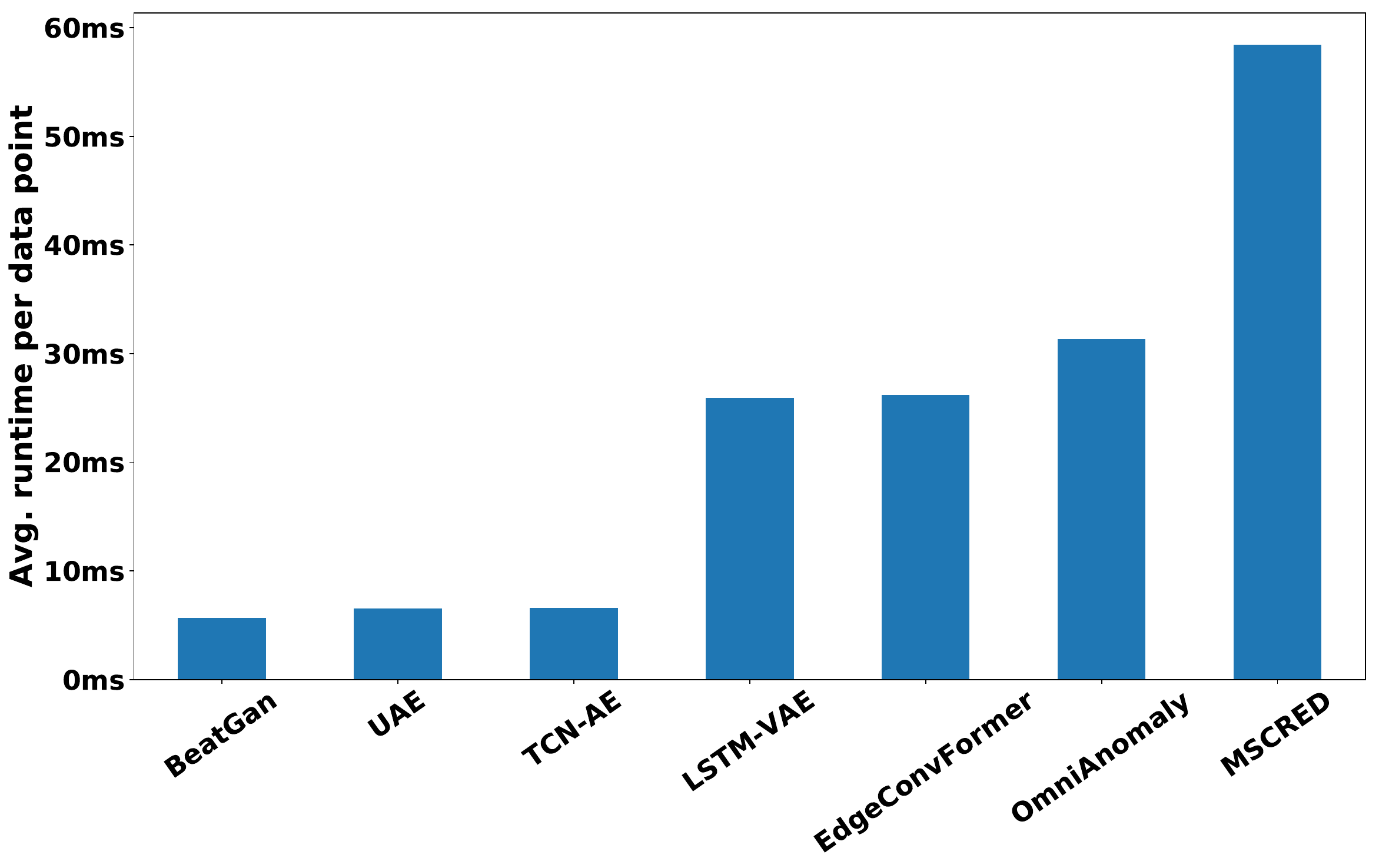}
\caption{Average runtime per algorithm for one data point}
\label{fig:runtime}
\end{figure}

\textbf{\textit{Baselines}}
Two sets of comparisons are conducted in our experiments. One is to compare with the top-performing algorithms described in \cite{garg2021evaluation}: LSTM-VAE \cite{park2018multimodal}, MSCRED \cite{zhang2019deep}, OmniAnomaly \cite{su2019robust}, BeatGAN \cite{zhou2019beatgan}, TCN-AE \cite{bai2018empirical}, UAE \cite{garg2021evaluation}. The other is to compare two recent transformer-based approaches: Anomaly Transformer \cite{xu2021anomaly} and TranAD\cite{tuli2022tranad}.

\begin{table}[ht]
\centering
\caption{Key hyperparameter values for each dataset in EdgeConvFormer.}

\begin{tabular}{ccccc}
   \toprule
   Dataset & Epochs & LR & Batch size & k value of KNN   \\
   \midrule
   SMD & 3  & 1e-3 & 32 & 8  \\
   MSL & 3 & 1e-3 & 32 & 4 \\
   SMAP & 3 & 1e-4 & 32 & 4 \\
   SWaT & 3 & 1e-2 & 32 & 8 \\
   PSM & 20 & 1e-3 & 32 & 6 \\
   Exathlon & 3 & 1e-3 & 32 &  varies from applications \\
   \bottomrule
\end{tabular}

\label{Tab:para}
\end{table}
\begin{table*}[hb]
\centering
\setlength\tabcolsep{1.5pt}
\caption{Key hyperparameter values used for each baseline.}
\begin{tabular}{ccccc}
   \toprule
   Model & Epochs & LR & Batch size & Design   \\
   \midrule
   UAE & 100  & 1e-3 & 256 & p=5  \\
   TCN-AE & 100 & 1.5 × 1e-4 & 128 & dropout=0.42, p=3, hidden-layers=min(10, int(m/6)), kernel-size=50 \\
   BeatGan & 100 & 1e-4 & 128 & z-dim=10, beta1=0.5  \\
   MSCRED & 100 & 1e-4 & 128 & step$_{max}$=5, s=3, w=[10,30,60] \\
   LSTM VAE & 100 & 9.5 × 1e-3 & 128 & Hidden layers=2, hidden-dim=15, z-dim=3, $\lambda_{reg}$ =0.55, $\lambda_{kulback}$=0.28 \\
   OmniAnomaly & 20 & 1e-3 & 50 & z-dim=3, RNN-units=500, dense-units=500, NF-layers=20 \\
   \bottomrule

\end{tabular}
\label{Tab:baseline}
\end{table*}

\subsection{Experimental result analysis}
\begin{table*}[ht]
\centering
\caption{Performance comparison of EdgeConvFormer with UAE, MSCRED, OmniAnomaly, BeatGan, LSTM-VAE, TCN-AE by using dynamic Gaussian scoring on the five public datasets. The best results are highlighted in bold.}

\resizebox{\linewidth}{!}{
\begin{tabular}{@{}cc|lll|ccc|ccc|ccc|ccc|c@{}}
\toprule
\multicolumn{2}{c|}{\textbf{Metric}}                                          & \multicolumn{3}{c|}{\textbf{F$_1$}}                                                                                  & \multicolumn{3}{c|}{\textbf{Fpa$_1$}}                                                                                & \multicolumn{3}{c|}{\textbf{Fc$_1$}}                                                                                 & \multicolumn{3}{c|}{\textbf{AU-ROC}} & \multicolumn{3}{c|}{\textbf{AU-PRC}} & \multicolumn{1}{c}{\textbf{\makecell[c]{Avg\\Rank}}}\\ \cmidrule(r){1-12}
\multicolumn{2}{c|}{\textbf{Thresholding}}                                    & \multicolumn{1}{c}{\textbf{Top-K}}  & \multicolumn{1}{c}{\textbf{Best-F$_1$}} & \multicolumn{1}{c|}{\textbf{Tail-P}} & \multicolumn{1}{c}{\textbf{Top-K}}  & \multicolumn{1}{c}{\textbf{Best-Fpa$_1$}} & \textbf{Tail-P}                      & \multicolumn{1}{c}{\textbf{Top-K}}  & \multicolumn{1}{c}{\textbf{Best-Fc$_1$}} & \multicolumn{1}{c|}{\textbf{Tail-P}} & \textbf{}  & \textbf{}  & \textbf{}  & \textbf{}  & \textbf{}  & \textbf{} \\ \midrule
\multicolumn{1}{c|}{\multirow{6}{*}{\textbf{SMD}}}  & \textbf{OmniAnomaly}            &0.4211 	&0.3519 	&0.2433
                               &\textbf{0.7670} 	&0.9548	&0.4471
                               & 0.5429 	&0.8233 	&0.3885
                               & \multicolumn{3}{c|}{0.8067 }          & \multicolumn{3}{c|}{0.4166}  & \multicolumn{1}{c}{4.1}         \\
\multicolumn{1}{c|}{}
& \textbf{LSTM-VAE}            & 0.3977                              & 0.2979                               & 0.3149                               &0.7321                                      & 0.9546                                     &0.8772                                     & 0.5241
& 0.7844              & \textbf{0.6213}
                               & \multicolumn{3}{c|}{0.7565}          & \multicolumn{3}{c|}{0.3816} & \multicolumn{1}{c}{4.9}         \\
\multicolumn{1}{c|}{}

& \textbf{MSCRED}            & 0.4111 	&\textbf{0.3537} 	&0.2653
                               &0.6484                                    & 0.8068                                      &0.1623                                    &0.3989 	& 0.5197 	&0.2711       & \multicolumn{3}{c|}{0.8103}          & \multicolumn{3}{c|}{0.4174}& \multicolumn{1}{c}{5.5}          \\
\multicolumn{1}{c|}{}
& \textbf{BeatGan}            &0.4056 	&0.2664 	&0.2924
                               &0.7539 	&0.9592 	&0.6940
                                  &0.5387 	&0.8069 	&0.4372     & \multicolumn{3}{c|}{0.8282}          & \multicolumn{3}{c|}{0.4014}& \multicolumn{1}{c}{4.7}          \\
\multicolumn{1}{c|}{}
& \textbf{TCN-AE}            &\textbf{0.4477} 	&0.2847 	&\textbf{0.3625}
                               & 0.7628 	&0.9810 	&\textbf{0.8893}
                               & \textbf{0.5818} 	&\textbf{0.8597} 	&0.5823
                               & \multicolumn{3}{c|}{0.8153 }          & \multicolumn{3}{c|}{\textbf{0.4268}}& \multicolumn{1}{c}{\textbf{2.0}}          \\
\multicolumn{1}{c|}{}
& \textbf{UAE}            & 0.4131                              & 0.2379                               & 0.3107                              & \multicolumn{1}{r}{0.7545}                & {0.9815}                 & \multicolumn{1}{r|}{0.8395}                & \multicolumn{1}{r}{0.5590}          & \multicolumn{1}{c}{0.8459}           & \multicolumn{1}{r|}{0.5944}          & \multicolumn{3}{c|}{\textbf{0.8283}} & \multicolumn{3}{c|}{0.3955} & \multicolumn{1}{c}{3.6}         \\
\multicolumn{1}{c|}{}                               & \textbf{EdgeConvFormer} & 0.4163                     & 0.2728                      &0.3276                      & \multicolumn{1}{c}{0.7546}          & \textbf{0.9816}                               & \multicolumn{1}{l|}{0.8678}          & \multicolumn{1}{c}{0.5564} & 0.8595                      & 0.6068                      & \multicolumn{3}{c|}{0.8039}          & \multicolumn{3}{c|}{0.3967} & \multicolumn{1}{c}{3.2}\\ \midrule
\multicolumn{1}{c|}{\multirow{6}{*}{\textbf{MSL}}}  & \textbf{OmniAnomaly}            &0.2308 	&0.1845 	&0.0824

                               &0.7319 	&0.9174 	&\textbf{0.5251}
                               &0.3330 	&0.5231 	&0.2823
                               & \multicolumn{3}{c|}{0.6476}          & \multicolumn{3}{c|}{0.2201}  & \multicolumn{1}{c}{3.8}        \\
\multicolumn{1}{c|}{}
& \textbf{LSTM-VAE}            & 0.2192 	&0.0973 	&0.0335
                              &0.7290                                 &0.9467                                       &0.1899                                      & 0.3178 		
                             & 0.6106                                & 0.0352                               & \multicolumn{3}{c|}{0.5841}          & \multicolumn{3}{c|}{0.2166} & \multicolumn{1}{c}{5.5}         \\
\multicolumn{1}{c|}{}

& \textbf{MSCRED}            &0.2332 	&0.2159 	&0.0339
                               & 0.7305 	&0.7696 	&0.0454
                                     & 0.3433 	&0.5055 	&0.0352
                               & \multicolumn{3}{c|}{0.6225}          & \multicolumn{3}{c|}{0.2556} & \multicolumn{1}{c}{4.5}         \\
\multicolumn{1}{c|}{}
& \textbf{BeatGan}            &0.2737 	&0.1421 	&0.1940
                               &0.7330 	&0.9359 	&0.1901
                               &0.3857 	&0.5570 	&0.1940
                                     & \multicolumn{3}{c|}{\textbf{0.6515}}          & \multicolumn{3}{c|}{0.2516}& \multicolumn{1}{c}{3.2}          \\
\multicolumn{1}{c|}{}
& \textbf{TCN-AE}            & 0.2561 	&0.1295 	&0.0339
                               & 0.7426 	&0.9642 	&0.0454
                               & 0.3645 	&0.6608 	&0.0352
                               & \multicolumn{3}{c|}{0.6057}          & \multicolumn{3}{c|}{0.2513}& \multicolumn{1}{c}{3.7}          \\
\multicolumn{1}{c|}{}
& \textbf{UAE}            & 0.1831                              & 0.0638                               & 0.1941                               & 0.7156                              & 0.9592                               & 0.1903                      & 0.2925                              & 0.4596                               & 0.1941                               & \multicolumn{3}{c|}{0.6178}          & \multicolumn{3}{c|}{0.1741} & \multicolumn{1}{c}{5.2}         \\
\multicolumn{1}{c|}{}                               & \textbf{EdgeConvFormer} & \multicolumn{1}{r}{\textbf{0.3307}} & \textbf{0.3404}                      & \textbf{0.2280}                      & \multicolumn{1}{c}{\textbf{0.7364}} & \textbf{0.9669}                      & \multicolumn{1}{l|}{0.1928}          & \textbf{0.4329}                     & \textbf{0.7520}                      & \textbf{0.4653}                      & \multicolumn{3}{c|}{0.5997} & \multicolumn{3}{c|}{\textbf{0.3247}}& \multicolumn{1}{c}{\textbf{1.6}} \\ \midrule
\multicolumn{1}{c|}{\multirow{6}{*}{\textbf{SMAP}}} & \textbf{OmniAnomaly}            &0.2272 	&\textbf{0.1881} 	&0.1406

                               & 0.7896	&0.9494	&0.2193
                               &0.2968 	&0.5881 	&0.1934
                               & \multicolumn{3}{c|}{0.6136}          & \multicolumn{3}{c|}{0.2269} & \multicolumn{1}{c}{4.2}         \\
\multicolumn{1}{c|}{}
& \textbf{LSTM-VAE}            & 0.1391                & 0.0646               & 0.1934
&0.7405 	&0.9783 	&0.2272
& 0.2165                              & 0.4868                               & 0.1934                               & \multicolumn{3}{c|}{0.5353}          & \multicolumn{3}{c|}{0.1382} & \multicolumn{1}{c}{5.6}         \\
\multicolumn{1}{c|}{}

& \textbf{MSCRED}            &0.2423 	&0.1487 	&0.1933
                                & 0.7640                                   & 0.8365                                     &\textbf{0.2297}                                   &0.3347 	       &0.5921 	     &0.1933
      & \multicolumn{3}{c|}{0.6168}          & \multicolumn{3}{c|}{0.2449} & \multicolumn{1}{c}{3.3}         \\
\multicolumn{1}{c|}{}
& \textbf{BeatGan}            &0.2393 	&0.1106 	&0.1928
                               &0.7664                                    &0.9796                                    &0.2280
                               &0.3318 	&0.6243 	&0.1928
      & \multicolumn{3}{c|}{0.6124}          & \multicolumn{3}{c|}{0.2267}& \multicolumn{1}{c}{4.1}          \\
\multicolumn{1}{c|}{}
& \textbf{TCN-AE}            & 0.2396 	&0.1861 	&0.1927
                               & 0.7889 	&\textbf{0.9845} 	&0.2272
                               & 0.3237 	&0.5861 	&0.1927
                               & \multicolumn{3}{c|}{0.6233}          & \multicolumn{3}{c|}{0.2208}  & \multicolumn{1}{c}{3.9}        \\
\multicolumn{1}{c|}{}
& \textbf{UAE}            & 0.1678                              & 0.1475                               & 0.1929                               & 0.7420                              & 0.9728                               & 0.2276                               & 0.2530                              & 0.5107                               & 0.1929                               & \multicolumn{3}{c|}{0.5678}          & \multicolumn{3}{c|}{0.1590} & \multicolumn{1}{c}{5.4}         \\
\multicolumn{1}{c|}{}                               & \textbf{EdgeConvFormer} & \textbf{0.3490}                     & 0.1766                      & \textbf{0.1942}                      & \multicolumn{1}{c}{\textbf{0.7912}} & \textbf{0.9845}  & \multicolumn{1}{l|}{0.2294} & \textbf{0.4759}                     & \textbf{0.6877}                      & \textbf{0.1942}                      & \multicolumn{3}{c|}{\textbf{0.7220}} & \multicolumn{3}{c|}{\textbf{0.3124}}& \multicolumn{1}{c}{\textbf{1.3}} \\ \midrule
\multicolumn{1}{c|}{\multirow{6}{*}{\textbf{SWAT}}} & \textbf{OmniAnomaly}            &0.2154 	&0.2425 	&0.2259

                               &0.5505 	&0.6358	&0.5707
                               &0.3256 	&0.4437 	&0.3502
                               & \multicolumn{3}{c|}{0.5386 }          & \multicolumn{3}{c|}{0.2135} & \multicolumn{1}{c}{6.5}         \\
\multicolumn{1}{c|}{}
& \textbf{LSTM-VAE}            &0.4008                               &0.3693                               &0.4036                                &0.5501                                     &0.6685                                       &0.5583                                     &0.4752
&0.5876              &0.4930
                               & \multicolumn{3}{c|}{0.6829}          & \multicolumn{3}{c|}{0.3417}& \multicolumn{1}{c}{4.5}          \\
\multicolumn{1}{c|}{}

& \textbf{MSCRED}            & 0.2324 	&0.1319 	&0.2691
                               &0.6945                                    &0.8359                                      &0.8168                                    &0.3572 	&0.5436 	&0.4033
 & \multicolumn{3}{c|}{0.6558}          & \multicolumn{3}{c|}{0.2298}  & \multicolumn{1}{c}{5.3}        \\
\multicolumn{1}{c|}{}
& \textbf{BeatGan}            &0.4901 	&0.2655 	&0.3652
                               &0.7523   &0.9006   &0.8926
                               &0.5905 	&0.6645 	&0.3645
                               & \multicolumn{3}{c|}{0.8569}          & \multicolumn{3}{c|}{0.5049}& \multicolumn{1}{c}{2.8}          \\
\multicolumn{1}{c|}{}
& \textbf{TCN-AE}            & 0.3315 	&0.2541 	&0.3399
                               &0.6926                                     & 0.8884                                      &0.8757                                     & 0.3315 	&0.2541 	&0.3399
      & \multicolumn{3}{c|}{0.7583}          & \multicolumn{3}{c|}{0.3462} & \multicolumn{1}{c}{4.8}         \\
\multicolumn{1}{c|}{}
& \textbf{UAE}            & 0.4735 	&0.0834 	&0.4683
                              & 0.7821                                    &\textbf{0.9616}                                      & \textbf{0.9464}                                     & 0.6239 		
                              & \textbf{0.7809}                      & 0.6690                               & \multicolumn{3}{c|}{0.8441
}          & \multicolumn{3}{c|}{0.4344} & \multicolumn{1}{c}{2.5}         \\
\multicolumn{1}{c|}{}                               & \textbf{EdgeConvFormer} & \textbf{0.7412}                     & \textbf{0.7275}                      & \textbf{0.7325}                     &\textbf{0.8209}                                     & {0.8833}                 &0.8703                                      & \textbf{0.6966}                     & 0.7258                               & \textbf{0.7169}                      & \multicolumn{3}{c|}{\textbf{0.9045}} & \multicolumn{3}{c|}{\textbf{0.7068}}& \multicolumn{1}{c}{\textbf{1.6}} \\ \midrule
\multicolumn{1}{c|}{\multirow{6}{*}{\textbf{PSM}}}  & \textbf{OmniAnomaly}            &0.4315 	&0.4106 	&0.3575

                               &0.7694 	&0.9300	&0.9259
                               &0.5479 	&0.5872 	&0.5775
                               & \multicolumn{3}{c|}{0.6462}          & \multicolumn{3}{c|}{0.4623}& \multicolumn{1}{c}{4.5}          \\
\multicolumn{1}{c|}{}
& \textbf{LSTM-VAE}            & 0.4216 	&0.2132 	&0.4421
                               &0.7749       	&0.9276       	&0.9211
                                     & 0.5599 	&0.6088 	&0.5948                               & \multicolumn{3}{c|}{0.6431}
                                     & \multicolumn{3}{c|}{0.4291}& \multicolumn{1}{c}{4.7}          \\
\multicolumn{1}{c|}{}

& \textbf{MSCRED}            & 0.4399                              & \textbf{0.5353}                               & 0.4688                              & 0.7486                                    & 0.8028                                      & 0.7903                                     & 0.4489                              & 0.5213                               & 0.4645                               & \multicolumn{3}{c|}{0.7172}          & \multicolumn{3}{c|}{0.4303}& \multicolumn{1}{c}{5.0}          \\
\multicolumn{1}{c|}{}
& \textbf{BeatGan}            &0.3579 	&0.0552 	&0.4075
                               & 0.7561                                   &0.9641                                   &0.9596
                               &0.5007 	&0.6948 	&0.6941
                                      & \multicolumn{3}{c|}{0.5893}          & \multicolumn{3}{c|}{0.3819}& \multicolumn{1}{c}{4.7}          \\
\multicolumn{1}{c|}{}
& \textbf{TCN-AE}            & \textbf{0.4965} 	&0.3041 	&\textbf{0.5092}
                               &\textbf{0.7981} 	&0.9402 	&0.9344
                                    &\textbf{0.6261} 	       &0.6840 	    &0.6801                               & \multicolumn{3}{c|}{\textbf{0.7268 }}          & \multicolumn{3}{c|}{0.5132} & \multicolumn{1}{c}{2.5}         \\
\multicolumn{1}{c|}{}
& \textbf{UAE}            & 0.3627                              & 0.0413                               & 0.3942                               & 0.7810                                    & \textbf{0.9741}                                      & 0.9650                                     & 0.5128                              & 0.6863                               & 0.6846                               & \multicolumn{3}{c|}{0.5746}          & \multicolumn{3}{c|}{0.3649}  & \multicolumn{1}{c}{4.5}        \\
\multicolumn{1}{c|}{}                               & \textbf{EdgeConvFormer} & 0.4655                     & 0.3265                      & 0.4803                      & 0.7592                     & 0.9740           & \textbf{0.9701}                               & \multicolumn{1}{c}{0.6043} & \textbf{0.7504}                      & \textbf{0.7465}                      & \multicolumn{3}{c|}{0.6835} & \multicolumn{3}{c|}{\textbf{0.5161}} & \multicolumn{1}{c}{\textbf{2.1}}         \\

\bottomrule
\end{tabular}
}
\label{Tab:suv}

\end{table*}

\textbf{\textit{Comparison with algorithms in \cite{garg2021evaluation}}}
As shown in TABLE~\ref{Tab:suv}, we compare the EdgeConvFormer model with the best performing models in UAE \cite{garg2021evaluation} and other top-ranked models - LSTM-VAE, MSCRED, BeatGan, TCN-AE, and OmniAnomaly, under the same scoring function (dynamic Gaussian scoring) with different threshold methods ($Best$-$F$-$score$, $Top$-$k$, and $Tail$-$p$), OmniAnomaly which has its own pre-defined scoring function. The results show that generally, the F$_1$ score is low, the Fpa$_1$ score is high, and the Fc$_1$ score is moderate. The proposed EdgeConvFormer model achieves the best performance -- top average rank under all threshold methods and evaluation metrics on MSL, SMAP, SWAT, and PSM datasets and the second-best performance on the SMD dataset. TABLE~\ref{Tab:rank} shows that in terms of the average rank across all five datasets, EdgeConvFormer outperforms all other algorithms. For the multi-entity datasets MSL and SMAP, we trained each entity separately. As shown in Table~\ref{Tab:dataset}, the average training data length of each entity in these two datasets is 2,160 and 2,556, with a relatively small amount of data. Although Transformer usually requires a large amount of training data, the proposed EdgeConvFormer model achieves the best anomaly detection performances in these two small datasets compared with other non-transformer algorithms, indicating that the EdgeConv module in our model has successfully played a compensating role. The ablation study in Table~\ref{Tab:ablation} also shows that removing the EdgeConv module leads to a significant performance drop.

According to the average rank in Table~\ref{Tab:rank}, the rank of algorithms are: EdgeConvFormer \begin{math} >\end{math} TCN-AE  \begin{math} >\end{math} BeatGan \begin{math}>\end{math} UAE \begin{math}>\end{math} OmniAnomaly \begin{math}>\end{math} MSCRED \begin{math}>\end{math}  LSTM-VAE. TCN-AE model performs well and even surpasses UAE when the restriction of using only the first sensor for anomaly detection is removed on the two datasets MSL and SMAP. This may be because TCN-AE is capable of learning long-term temporal patterns in the input data with a hierarchical temporal model established with dilated convolutions \cite{bai2018empirical}. BeatGan \cite{zhou2019beatgan} also performs well, ranking third, mainly thanks to the use of a Generative Adversarial Network (GAN) framework, where the reconstruction errors produced by the generator are regularized by the discriminator. Its performance is not optimal probably because the generator uses CNN as the encoder and decoder to reconstruct the time series, while the filter of CNN only slides along the time dimension, ignoring the relationship between sensors. The filter size is crucial to its performance and selecting an appropriate size to capture long-term dependencies requires careful tuning. The UAE model also performs well mainly because it consists of independent sensor-wise fully-connected autoencoder models that learn representation for each sensor separately before aggregating scores across all the sensors. This design enables UAE to effectively detect temporal anomalies, especially when the intra-sensor temporal association impacts more than the inter-sensor correlation. For example, the \begin{math}Fpa_{1}\end{math} score of UAE is much higher than other algorithms on the SWAT (Secure Water Treatment) dataset. The reason may be that SWAT consists of six sub-processes, in which signals are mixed up with different sub-processes from sensors and actuators. The actuators only include two statuses: open and closed, denoted as 1 and 0, respectively, and are not so informative as sensors \cite{lin2018tabor}. Hence the inter-sensor correlations are not that significant, while temporal correlations within each sensor are prevalent. LSTM-VAE and OmniAnomaly are RNN-based methods. The OmniAnomaly model performs better than LSTM-VAE because it not only glues together GRU (Gating recursive units) and VAE (variational automatic encoders) but also employs technics such as stochastic variable connection to model complex temporal dependence and stochasticity of multivariate time series. One limitation of OmniAnomaly is that the static scoring function cannot adapt to the changing norm during testing, nor anomalies in the test set \cite{su2019robust, garg2021evaluation}. The limited ability of the multi-scale signature matrices to characterize the state of the system and the possible loss of information may be the reasons for the poor performance of MSCRED.
\begin{table}[]

  \centering
  \begin{threeparttable}
  \caption{Average rank of the algorithms across all five datasets.}
  \label{tab:performance_comparison}

    \begin{tabular}{ccccccc}
    \toprule

   \textbf{Algorithms}&\textbf{\makecell[c]{Avg-rank \\on SMD}}&\textbf{\makecell[c]{Avg-rank \\on MSL}}&\textbf{\makecell[c]{Avg-rank \\on SMAP}}&\textbf{\makecell[c]{Avg-rank \\on SWAT}}&\textbf{\makecell[c]{Avg-rank \\on PSM}}&\textbf{\makecell[c]{overall\\Avg-rank}}\cr
    \midrule
    \textbf{OmniAnomaly}&4.1&3.8&4.2&6.5&4.5&4.62\cr
    \textbf{LSTM-VAE}&4.9&5.5&5.6&4.5&4.7&5.04\cr
    \textbf{MSCRED}&5.5&4.5&3.3&5.3&5.0&4.72\cr
    \textbf{BeatGan}&4.7&3.2&4.1&2.8&4.7&3.90\cr
    \textbf{TCN-AE}&\textbf{2.0}&3.7&3.9&4.8&2.5&3.38\cr
    \textbf{UAE}&3.6&5.2&5.4&2.5&4.5&4.24\cr
    \textbf{EdgeConvFormer(ours)}&3.2&\textbf{1.6}&\textbf{1.3}&\textbf{1.6}&\textbf{2.1}&\textbf{1.96}\cr
    \bottomrule
    \end{tabular} 
    
    \label{Tab:rank}
    \end{threeparttable}
\end{table}

\textbf{\textit{Comparison with the two Transformer-based approaches}}
As shown in TABLE~\ref{Tab:sota}, since the complete code of the Anomaly Transformer \cite{xu2021anomaly} is not available, we use the results reported in their paper to compare with the reproduced results of TranAD \cite{tuli2022tranad} and the experimental results of our EdgeConvFormer using the same evaluation metrics. TranAD \cite{tuli2022tranad} proposes an adversarial training procedure consisting of two Transformer encoders and two Transformer decoders to amplify reconstruction errors since simple Transformer-based networks tend to miss small anomalous deviations. Anomaly Transformer \cite{xu2021anomaly} has a similar motivation to TranAD: to make rare abnormalities easier to distinguish. The innovations of the Anomaly Transformer lie in its proposal of an anomaly transformer to model prior association and series association simultaneously and a minimax strategy to amplify the normal-abnormal distinguishability to detect anomalies. Both the TranAD and Anomaly Transformer only focus on time series from a single source and do not consider the interdependencies between multiple sensors, which limits their ability to detect and explain anomalies. When comparing these two models, experimental results show that, except for the SWAT dataset, the \begin{math}Fpa_{1}\end{math} scores of the EdgeConvFormer model on SMD, MSL, and SMAP outperform all the other baselines. For the PSM dataset, EdgeConvFormer achieves 97.40\% which is close to the best score - 97.89\%.

\begin{table*}
\centering
\caption{Performance comparison of EdgeConvFormer with SOTA on the five public datasets. P: Precision (as \%), R: Recall (as \%),  Fpa$_1$: point-adjusted F$_1$ score (as \%). The best results are highlighted in bold.}
 \resizebox{\linewidth}{!}{
\begin{tabular}{c|lll|lll|lll|lll|lll}

\hline
\textbf{Dateset}              & \multicolumn{3}{c|}{\textbf{SMD}}                                                                    & \multicolumn{3}{c|}{\textbf{MSL}}                                                                    & \multicolumn{3}{c|}{\textbf{SMAP}}                                                                   & \multicolumn{3}{c|}{\textbf{SWAT}}                                                                   & \multicolumn{3}{c}{\textbf{PSM}}                                                                    \\
\textbf{Metric}               & \multicolumn{1}{c}{\textbf{P}} & \multicolumn{1}{c}{\textbf{R}} & \multicolumn{1}{c|}{\textbf{Fpa$_1$}} & \multicolumn{1}{c}{\textbf{P}} & \multicolumn{1}{c}{\textbf{R}} & \multicolumn{1}{c|}{\textbf{Fpa$_1$}} & \multicolumn{1}{c}{\textbf{P}} & \multicolumn{1}{c}{\textbf{R}} & \multicolumn{1}{c|}{\textbf{Fpa$_1$}} & \multicolumn{1}{c}{\textbf{P}} & \multicolumn{1}{c}{\textbf{R}} & \multicolumn{1}{c|}{\textbf{Fpa$_1$}} & \multicolumn{1}{c}{\textbf{P}} & \multicolumn{1}{c}{\textbf{R}} & \multicolumn{1}{c}{\textbf{Fpa$_1$}} \\ \hline
\textbf{LOF}                  & 56.34                          & 39.86                          & 46.68                              & 47.72                          & 85.25                          & 61.18                              & 58.93                          & 56.33                          & 57.60                               & 72.15                          & 65.43                          & 68.62                              & 57.89                          & 90.49                          & 70.61                             \\
\textbf{OCSVM}                & 44.34                          & 76.72                          & 56.19                              & 59.78                          & 86.87                          & 70.82                              & 53.85                          & 59.07                          & 56.34                              & 45.39                          & 49.22                          & 47.23                              & 62.75                          & 80.89                          & 70.67                             \\
\textbf{IsolationForest}      & 42.31                          & 73.29                          & 53.64                              & 53.94                          & 86.54                          & 66.45                              & 52.39                          & 59.07                          & 55.53                              & 49.29                          & 44.95                          & 47.02                              & 76.09                          & 92.45                          & 83.48                             \\
\textbf{Deep-SVDD}            & 78.54                          & 79.67                          & 79.10                               & 91.92                          & 76.63                          & 83.58                              & 89.93                          & 56.02                          & 69.04                              & 80.42                          & 84.45                          & 82.39                              & 95.41                          & 86.49                          & 90.73                             \\
\textbf{LSTM-VAE}             & 75.76                          & 90.08                          & 82.30                               & 85.49                          & 79.94                          & 82.62                              & 92.20                           & 67.75                          & 78.10                               & 76.00                             & 89.50                           & 82.20                               & 73.62                          & 89.92                          & 80.96                             \\
\textbf{MSCRED}               & 72.76                          & 99.74                          & 84.14                              & 89.12                          & 98.62                          & 93.63                              & 81.75                          & 92.16                          & 86.64                              & 99.92                          & 67.70                           & 80.72                              & 92.80                               &70.74                                & 80.28                                  \\
\textbf{DAGMM}                & 67.30                           & 49.89                          & 57.30                               & 89.6                           & 63.93                          & 74.62                              & 86.45                          & 56.73                          & 68.51                              & 89.92                          & 57.84                          & 70.40                              & 93.49                          & 70.03                          & 80.08                             \\
\textbf{OmniAnomaly}          & 83.68                          & 86.82                          & 85.22                              & 89.02                          & 86.37                          & 87.67                              & 92.49                          & 81.99                          & 86.92                              & 81.42                          & 84.30                           & 82.83                              & 88.39                          & 74.46                          & 80.83                             \\
\textbf{BeatGAN}              & 72.90                           & 84.09                          & 78.10                               & 89.75                          & 85.42                          & 87.53                              & 92.38                          & 55.85                          & 69.61                              & 64.01                          & 87.46                          & 73.92                              & 90.30                           & 93.84                          & 92.04                             \\
\textbf{InterFusion}          & 87.02                          & 85.43                          & 86.22                              & 81.28                          & 92.70                           & 86.62                              & 89.77                          & 88.52                          & 89.14                              & 80.59                          & 85.58                          & 83.01                              & 83.61                          & 83.45                          & 83.52                             \\
\textbf{THOC}                 & 79.76                          & 90.95                          & 84.99                              & 88.45                          & 90.97                          & 89.69                              & 92.06                          & 89.34                          & 90.68                              & 83.94                          & 86.36                          & 85.13                              & 88.14                          & 90.99                          & 89.54                             \\

\textbf{Anomaly Transformer}  & 89.40                           & 95.45                          & 92.33                              & 92.09                          & 95.15                          & 93.59                              & 94.13                          & 99.40                          & 96.69                              & 91.55                          & \textbf{96.73}                          & \textbf{94.07}                              & 96.91                          & \textbf{98.90}                  & \textbf{97.89}                    \\
\textbf{TranAD}               & 92.62                          & \textbf{99.74}                          & 96.05                              & 90.38                          & \textbf{99.99}                 & 94.94                              & 80.43                          & \textbf{99.99}                 & 89.15                              & 97.60                           & 69.97                          & 81.51                              & 68.13                          & 90.18                          & 77.62
                             \\

\hline
\textbf{EdgeConvFormer(ours)} & \textbf{97.83}                 & 98.50                 & \textbf{98.16}                     & \textbf{95.78}                 & 97.61                          & \textbf{96.69}                     & \textbf{97.33}                 & 99.61                 & \textbf{98.45}                     &\textbf{97.81}                                &80.52                                &88.33                                    & \textbf{98.13}                 & 96.68                          & 97.40                             \\ \hline
\end{tabular}
}
\label{Tab:sota}

\end{table*}

\textbf{\textit{Experiment on Exathlon dataset}}
In this dataset, we adopted new evaluation metrics: range-based $P_{T}$, $R_{T}$ and $F_{1}$-$P_{T}$$R_{T}$ at different AD levels (AD1-4), from basic to advanced, where a higher AD level includes the requirements of all preceding levels. So the higher the level, the higher the requirement, and the lower the score. We selected three Exathlon applications: application 1, application 2, and application 9, and carried out anomaly detection on these three applications, respectively. Application 1 containes three exception types: T2, T4 and T5; application 2 containes three types of anomalies: T1, T2, T5 (2 traces); application 9 contains four exception types: T2, T3, T4, and T5.  We ran each application, leading to different performance results for each anomaly type, for which we report the mean performance in Table ~\ref{Tab:ExathlonResults}. The experimental results show that: 1) For different anomaly types, the performance for T2 (Bursty Input Until Crash) and T1 (bursty input) are better than other anomaly types across all methods, showing that the variety of the anomaly types in this dataset present signals of different strength levels in the data. 2) On the four levels and almost all anomaly types, EdgeConvFormer's $F_{1}$-$P_{T}$$R_{T}$ scores outperform other algorithms. 3) With the increase in AD level, $F_{1}$-$P_{T}$$R_{T}$ scores of all algorithms decline. The proposed EdgeConvFormer has shown outstanding robustness for anomaly detection, i.e., it achieves satisfying performances in the detection of all anomaly types. For example, while most algorithms failed to detect any anomaly in AD4, EdgeConvFormer achieves $0.7309$ $F_{1}$-$P_{T}$$R_{T}$ score. The performance decrease of EdgeConvFormer in AD4 is much less compared to other algorithms, indicating that the detection ability of EdgeConvFormer not only catches the existence of an anomaly but also meets the following requirements: the larger size of the correctly predicted portion; the relative position of the correctly predicted part is in good agreement with the position where the actual anomaly started; detecting each anomaly segment with a single prediction range.

\begin{table*}[]
\caption{Anomaly detection results on Exathlon dataset.  AD1 (Anomaly Existence), AD2 (Range Detection), AD3 (Early Detection), and AD4 (Exactly-Once Detection). $P_{T}$: Range-based Precision (as \%), $R_{T}$: Range-based Recall (as \%),  $F_{1}$-$P_{T}$$R_{T}$: Range-based F$_1$ score (as \%). The best results are highlighted in bold.}

\resizebox{\linewidth}{!}{
\begin{tabular}{@{}c|ccccc|ccccc|ccccc@{}}
\toprule
\rowcolor[HTML]{EFEFEF}
\textbf{AD1} & \multicolumn{5}{c|}{\cellcolor[HTML]{EFEFEF}\textbf{\pmb{$P_{T}$} for Anomaly Types T1$\rightarrow$T5}}                                                                                                            & \multicolumn{5}{c|}{\cellcolor[HTML]{EFEFEF}\textbf{\pmb{$R_{T}$} for Anomaly Types T1$\rightarrow$T5}}                                                                                                                                & \multicolumn{5}{c|}{\cellcolor[HTML]{EFEFEF}\textbf{\pmb{$F_{1}$-$P_{T}$$R_{T}$} for Anomaly Types T1$\rightarrow$ T5}}                                                                                                       \\ \midrule
\textbf{OmniAnomaly}                                       & {\color[HTML]{000000} 0.4543}          & {\color[HTML]{000000} 0.4310}          & {\color[HTML]{000000} 0.3913}          & {\color[HTML]{000000} 0.3500}          & {\color[HTML]{330001} 0.3667}          & 1.0                                                        & 1.0                                    & {\color[HTML]{000000} 1.0}             & {\color[HTML]{000000} 0.3333}          & 1.0                                    & {\color[HTML]{000000} 0.6247}          & {\color[HTML]{000000} 0.6024}          & {\color[HTML]{000000} 0.5625}          & {\color[HTML]{000000} 0.3415}          & {\color[HTML]{000000} 0.5366}          \\
\textbf{LSTM-VAE}                                          & {\color[HTML]{000000} 0.9193}          & {\color[HTML]{000000} 0.9411}          & {\color[HTML]{000000} 0.5556}          & {\color[HTML]{000000} 0.1259}          & {\color[HTML]{330001} 0.6047}          & 1.0                                                        & 1.0                                    & {\color[HTML]{000000} 0.75}            & {\color[HTML]{000000} \textbf{1.0}}    & 1.0                                    & {\color[HTML]{000000} 0.9580}          & {\color[HTML]{000000} 0.9744}          & {\color[HTML]{000000} 0.6383}          & {\color[HTML]{000000} 0.2237}          & {\color[HTML]{000000} 0.7537}          \\
\textbf{MSCRED}                                            & {\color[HTML]{000000} 0.2973}          & {\color[HTML]{000000} 0.9167}          & {\color[HTML]{000000} 0.4802}          & {\color[HTML]{000000} \textbf{0.6347}} & {\color[HTML]{330001} 0.7925}          & 1.0                                                        & 1.0                                    & {\color[HTML]{000000} 1.0}             & {\color[HTML]{000000} 0.6667}          & 0.8                                    & {\color[HTML]{000000} 0.4583}          & {\color[HTML]{000000} 0.9565}          & {\color[HTML]{000000} 0.6488}          & {\color[HTML]{000000} 0.6503}          & {\color[HTML]{000000} 0.7962}          \\
\textbf{BeatGan}                                           & {\color[HTML]{000000} 0.9119}          & {\color[HTML]{000000} 0.9474}          & {\color[HTML]{000000} 0.2714}          & {\color[HTML]{000000} 0.3250}          & {\color[HTML]{330001} 0.6000}          & 1.0                                                        & 1.0                                    & {\color[HTML]{000000} 0.5}             & {\color[HTML]{000000} \textbf{1.0}}    & 0.8                                    & {\color[HTML]{000000} 0.9539}          & {\color[HTML]{000000} 0.9730}          & {\color[HTML]{000000} 0.3518}          & {\color[HTML]{000000} 0.4906}          & {\color[HTML]{000000} 0.6857}          \\
\textbf{TCN-AE}                                            & {\color[HTML]{000000} 0.5571}          & {\color[HTML]{000000} 0.8571}          & {\color[HTML]{000000} 0.3115}          & {\color[HTML]{000000} 0.2857}          & {\color[HTML]{330001} 0.7460}          & 1.0                                                        & 1.0                                    & {\color[HTML]{000000} 0.5}             & {\color[HTML]{000000} 0.3333}          & 1.0                                    & {\color[HTML]{000000} 0.7156}          & {\color[HTML]{000000} 0.9231}          & {\color[HTML]{000000} 0.3839}          & {\color[HTML]{000000} 0.3077}          & {\color[HTML]{000000} 0.8545}          \\
\textbf{UAE}                                               & {\color[HTML]{000000} 0.8776}          & {\color[HTML]{000000} 0.9697}          & {\color[HTML]{000000} 0.5000}          & {\color[HTML]{000000} 0.5000}          & {\color[HTML]{330001} \textbf{0.8699}} & 1.0                                                        & 1.0                                    & {\color[HTML]{000000} 0.25}            & {\color[HTML]{000000} 0.1667}          & 1.0                                    & {\color[HTML]{000000} 0.9348}          & {\color[HTML]{000000} 0.9846}          & {\color[HTML]{000000} 0.3333}          & {\color[HTML]{000000} 0.2500}          & {\color[HTML]{000000} \textbf{0.9304}}          \\
\textbf{EdgeConvFormer}                                    & {\color[HTML]{000000} \textbf{0.9347}} & {\color[HTML]{000000} \textbf{0.9847}} & {\color[HTML]{000000} \textbf{0.5667}} & {\color[HTML]{000000} 0.5870}          & {\color[HTML]{330001} 0.8349}          & \textbf{1.0}                                               & \textbf{1.0}                           & {\color[HTML]{000000} \textbf{1.0}}    & {\color[HTML]{000000} 0.8333}          & \textbf{1.0}                           & {\color[HTML]{000000} \textbf{0.9662}} & {\color[HTML]{000000} \textbf{0.9923}} & {\color[HTML]{000000} \textbf{0.7234}} & {\color[HTML]{000000} \textbf{0.6888}} & {\color[HTML]{000000} 0.9100}          \\ \midrule
\rowcolor[HTML]{DAE8FC}
\textbf{AD2} & \multicolumn{5}{c|}{\cellcolor[HTML]{DAE8FC}\textbf{\pmb{$P_{T}$} for Anomaly Types T1 $\rightarrow$ T5}}                                                                                                            & \multicolumn{5}{c|}{\cellcolor[HTML]{DAE8FC}\textbf{\pmb{$R_{T}$} for Anomaly Types T1 $\rightarrow$ T5}}                                                                                                                                & \multicolumn{5}{c}{\cellcolor[HTML]{DAE8FC}{\color[HTML]{000000} \textbf{\pmb{$F_{1}$-$P_{T}$$R_{T}$} for Anomaly Types T1 $\rightarrow$ T5}}}                                                                                \\ \midrule
\textbf{OmniAnomaly}                                       & {\color[HTML]{000000} \textbf{0.9313}} & {\color[HTML]{000000} \textbf{1.0}}    & {\color[HTML]{000000} 0.2034}          & {\color[HTML]{000000} 0.2439}          & {\color[HTML]{000000} 0.2394}          & \multicolumn{1}{l}{{\color[HTML]{000000} 0.2902}}          & {\color[HTML]{000000} 0.0120}          & {\color[HTML]{000000} 0.0031}          & {\color[HTML]{000000} 0.0493}          & {\color[HTML]{000000} 0.0154}          & {\color[HTML]{000000} 0.4425}          & {\color[HTML]{000000} 0.0238}          & {\color[HTML]{000000} 0.0061}          & {\color[HTML]{000000} 0.0820}          & {\color[HTML]{000000} 0.0290}          \\
\textbf{LSTM-VAE}                                          & {\color[HTML]{000000} 0.8125}          & {\color[HTML]{000000} 0.3755}          & {\color[HTML]{000000} 0.5556}          & {\color[HTML]{000000} 0.1259}          & {\color[HTML]{000000} 0.7248}          & \multicolumn{1}{l}{{\color[HTML]{000000} 0.7827}}          & {\color[HTML]{000000} 0.6889}          & {\color[HTML]{000000} 0.2608}          & {\color[HTML]{000000} 0.3365}          & {\color[HTML]{000000} 0.4100}          & {\color[HTML]{000000} 0.7973}          & {\color[HTML]{000000} 0.4860}          & {\color[HTML]{000000} 0.3549}          & {\color[HTML]{000000} 0.1833}          & {\color[HTML]{000000} 0.5237}          \\
\textbf{MSCRED}                                            & {\color[HTML]{000000} 0.4268}          & {\color[HTML]{000000} 0.5890}          & {\color[HTML]{000000} 0.3076}          & {\color[HTML]{000000} 0.6667}          & {\color[HTML]{000000} 0.5261}          & \multicolumn{1}{l}{{\color[HTML]{000000} \textbf{0.9316}}} & {\color[HTML]{000000} 0.9801}          & {\color[HTML]{000000} 0.8698}          & {\color[HTML]{000000} 0.3884}          & {\color[HTML]{000000} 0.3148}          & {\color[HTML]{000000} 0.5854}          & {\color[HTML]{000000} 0.7358}          & {\color[HTML]{000000} 0.4546}          & {\color[HTML]{000000} 0.4908}          & {\color[HTML]{000000} 0.3939}          \\
\textbf{BeatGan}                                           & {\color[HTML]{000000} 0.7894}          & {\color[HTML]{000000} 0.8793}          & {\color[HTML]{000000} 0.2498}          & {\color[HTML]{000000} 0.3250}          & {\color[HTML]{000000} 0.4233}          & \multicolumn{1}{l}{{\color[HTML]{000000} 0.6120}}          & {\color[HTML]{000000} 0.9115}          & {\color[HTML]{000000} 0.4611}          & {\color[HTML]{000000} \textbf{0.7907}} & {\color[HTML]{000000} 0.1868}          & {\color[HTML]{000000} 0.6895}          & {\color[HTML]{000000} 0.8951}          & {\color[HTML]{000000} 0.3241}          & {\color[HTML]{000000} 0.4606}          & {\color[HTML]{000000} 0.2593}          \\
\textbf{TCN-AE}                                            & {\color[HTML]{000000} 0.6667}          & {\color[HTML]{000000} 0.8166}          & {\color[HTML]{000000} 0.2185}          & {\color[HTML]{000000} 0.1197}          & {\color[HTML]{000000} 0.5795}          & \multicolumn{1}{l}{{\color[HTML]{000000} 0.8650}}          & {\color[HTML]{000000} 0.9687}          & {\color[HTML]{000000} 0.3791}          & {\color[HTML]{000000} 0.7521}          & {\color[HTML]{000000} 0.2424}          & {\color[HTML]{000000} 0.7530}          & {\color[HTML]{000000} 0.8862}          & {\color[HTML]{000000} 0.2772}          & {\color[HTML]{000000} 0.2065}          & {\color[HTML]{000000} 0.3418}          \\
\textbf{UAE}                                               & {\color[HTML]{000000} 0.7844}          & {\color[HTML]{000000} 0.6154}          & {\color[HTML]{000000} 0.2961}          & {\color[HTML]{000000} 0.0915}          & {\color[HTML]{000000} \textbf{0.8610}} & \multicolumn{1}{l}{{\color[HTML]{000000} 0.5964}}          & {\color[HTML]{000000} 0.9151}          & {\color[HTML]{000000} 0.3207}          & {\color[HTML]{000000} 0.7550}          & {\color[HTML]{000000} 0.3261}          & {\color[HTML]{000000} 0.6776}          & {\color[HTML]{000000} 0.7369}          & {\color[HTML]{000000} 0.3079}          & {\color[HTML]{000000} 0.1633}          & {\color[HTML]{000000} 0.4730}          \\
\textbf{EdgeConvFormer}                                    & {\color[HTML]{000000} 0.7600}          & {\color[HTML]{000000} 0.8904}          & {\color[HTML]{000000} \textbf{0.5621}} & {\color[HTML]{000000} \textbf{0.7622}} & {\color[HTML]{000000} 0.4551}          & \multicolumn{1}{l}{{\color[HTML]{000000} 0.8529}}          & {\color[HTML]{000000} \textbf{0.9898}} & {\color[HTML]{000000} \textbf{0.9170}} & {\color[HTML]{000000} 0.5744}          & {\color[HTML]{000000} \textbf{0.9549}} & {\color[HTML]{000000} \textbf{0.8038}} & {\color[HTML]{000000} \textbf{0.9375}} & {\color[HTML]{000000} \textbf{0.6969}} & {\color[HTML]{000000} \textbf{0.6551}} & {\color[HTML]{000000} \textbf{0.6164}} \\ \midrule
\rowcolor[HTML]{34CDF9}
\textbf{AD3} & \multicolumn{5}{c|}{\cellcolor[HTML]{34CDF9}\textbf{\pmb{$P_{T}$} for Anomaly Types T1 $\rightarrow$ T5}}                                                                                                            & \multicolumn{5}{c|}{\cellcolor[HTML]{34CDF9}\textbf{\pmb{$R_{T}$} for Anomaly Types T1 $\rightarrow$ T5}}                                                                                                                                & \multicolumn{5}{c|}{\cellcolor[HTML]{34CDF9}{\color[HTML]{000000} \textbf{\pmb{$F_{1}$-$P_{T}$$R_{T}$} for Anomaly Types T1 $\rightarrow$ T5}}}                                                                                \\ \midrule
\textbf{OmniAnomaly}                                       & {\color[HTML]{000000} \textbf{0.9285}} & {\color[HTML]{000000} 0.0008}          & {\color[HTML]{000000} 0.2034}          & {\color[HTML]{000000} 0.2439}          & {\color[HTML]{000000} 0.2394}          & {\color[HTML]{000000} 0.1821}                              & {\color[HTML]{000000} 0.0016}          & {\color[HTML]{000000} 0.0021}          & {\color[HTML]{000000} 0.0151}          & {\color[HTML]{000000} 0.0125}          & {\color[HTML]{000000} 0.3045}          & {\color[HTML]{000000} 0.0102}          & {\color[HTML]{000000} 0.0042}          & {\color[HTML]{000000} 0.0285}          & {\color[HTML]{000000} 0.0237}          \\
\textbf{LSTM-VAE}                                          & {\color[HTML]{000000} 0.8125}          & {\color[HTML]{000000} \textbf{0.9929}} & {\color[HTML]{000000} 0.5556}          & {\color[HTML]{000000} 0.1259}          & {\color[HTML]{000000} 0.3810}          & {\color[HTML]{000000} 0.5448}                              & {\color[HTML]{000000} 0.3266}          & {\color[HTML]{000000} 0.2545}          & {\color[HTML]{000000} 0.2455}          & {\color[HTML]{000000} 0.5349}          & {\color[HTML]{000000} 0.6523}          & {\color[HTML]{000000} 0.4915}          & {\color[HTML]{000000} 0.3491}          & {\color[HTML]{000000} 0.1665}          & {\color[HTML]{000000} 0.4450}          \\
\textbf{MSCRED}                                            & {\color[HTML]{000000} 0.3733}          & {\color[HTML]{000000} 0.5890}          & {\color[HTML]{000000} 0.3077}          & {\color[HTML]{000000} 0.5000}          & {\color[HTML]{000000} 0.5261}          & {\color[HTML]{000000} \textbf{0.9277}}                     & {\color[HTML]{000000} \textbf{0.9801}} & {\color[HTML]{000000} 0.7608}          & {\color[HTML]{000000} 0.3670}          & {\color[HTML]{000000} 0.2943}          & {\color[HTML]{000000} 0.5324}          & {\color[HTML]{000000} 0.7358}          & {\color[HTML]{000000} 0.4382}          & {\color[HTML]{000000} 0.4233}          & {\color[HTML]{000000} 0.3775}          \\
\textbf{BeatGan}                                           & {\color[HTML]{000000} 0.7894}          & {\color[HTML]{000000} 0.8793}          & {\color[HTML]{000000} 0.2499}          & {\color[HTML]{000000} 0.3250}          & {\color[HTML]{000000} 0.4233}          & {\color[HTML]{000000} 0.4521}                              & {\color[HTML]{000000} 0.7520}          & {\color[HTML]{000000} 0.3964}          & {\color[HTML]{000000} \textbf{0.7303}} & {\color[HTML]{000000} 0.1732}          & {\color[HTML]{000000} 0.5749}          & {\color[HTML]{000000} 0.8106}          & {\color[HTML]{000000} 0.3065}          & {\color[HTML]{000000} 0.4498}          & {\color[HTML]{000000} 0.2487}          \\
\textbf{TCN-AE}                                            & {\color[HTML]{000000} 0.6667}          & {\color[HTML]{000000} 0..8166}         & {\color[HTML]{000000} 0.2185}          & {\color[HTML]{000000} 0.1197}          & {\color[HTML]{000000} 0.2277}          & {\color[HTML]{000000} 0.5825}                              & {\color[HTML]{000000} 0.6360}          & {\color[HTML]{000000} 0.3437}          & {\color[HTML]{000000} 0.6753}          & {\color[HTML]{000000} 0.4868}          & {\color[HTML]{000000} 0.6218}          & {\color[HTML]{000000} 0.7151}          & {\color[HTML]{000000} 0.2672}          & {\color[HTML]{000000} 0.2033}          & {\color[HTML]{000000} 0.3103}          \\
\textbf{UAE}                                               & {\color[HTML]{000000} 0.4984}          & {\color[HTML]{000000} 0.6154}          & {\color[HTML]{000000} 0.2961}          & {\color[HTML]{000000} 0.0915}          & {\color[HTML]{000000} \textbf{0.8610}} & {\color[HTML]{000000} 0.9183}                              & {\color[HTML]{000000} 0.5642}          & {\color[HTML]{000000} 0.2891}          & {\color[HTML]{000000} 0.7110}          & {\color[HTML]{000000} 0.2901}          & {\color[HTML]{000000} 0.6461}          & {\color[HTML]{000000} 0.5887}          & {\color[HTML]{000000} 0.2925}          & {\color[HTML]{000000} 0.1622}          & {\color[HTML]{000000} 0.4340}          \\
\textbf{EdgeConvFormer}                                    & {\color[HTML]{000000} 0.5751}          & {\color[HTML]{000000} 0.8904}          & {\color[HTML]{000000} \textbf{0.5621}} & {\color[HTML]{000000} \textbf{0.6097}} & {\color[HTML]{000000} 0.4551}          & {\color[HTML]{000000} 0.9229}                              & {\color[HTML]{000000} 0.8813}          & {\color[HTML]{000000} \textbf{0.8613}} & {\color[HTML]{000000} 0.5744}          & {\color[HTML]{000000} \textbf{0.8952}} & {\color[HTML]{000000} \textbf{0.7086}} & {\color[HTML]{000000} \textbf{0.8858}} & {\color[HTML]{000000} \textbf{0.6802}} & {\color[HTML]{000000} \textbf{0.5915}} & {\color[HTML]{000000} \textbf{0.6034}} \\ \midrule
\rowcolor[HTML]{3166FF}
\textbf{AD4} & \multicolumn{5}{c|}{\cellcolor[HTML]{3166FF}\textbf{\pmb{$P_{T}$} for Anomaly Types T1 $\rightarrow$ T5}}                                                                                                            & \multicolumn{5}{c|}{\cellcolor[HTML]{3166FF}\textbf{\pmb{$R_{T}$} for Anomaly Types T1 $\rightarrow$ T5}}                                                                                                                                & \multicolumn{5}{c|}{\cellcolor[HTML]{3166FF}{\color[HTML]{000000} \textbf{\pmb{$F_{1}$-$P_{T}$$R_{T}$} for Anomaly Types T1 $\rightarrow$ T5}}}                                                                                \\ \midrule
\textbf{OmniAnomaly}                                       & {\color[HTML]{330001} \textbf{0.9445}} & {\color[HTML]{000000} \textbf{1.0}}    & {\color[HTML]{000000} 0.2619}          & {\color[HTML]{000000} 0.2439}          & {\color[HTML]{343434} 0.3667}          & {\color[HTML]{000000} 0}                                   & {\color[HTML]{000000} 0}               & {\color[HTML]{000000} 0}               & {\color[HTML]{000000} 0.0002}          & {\color[HTML]{000000} 0.0059}          & {\color[HTML]{000000} 0}               & {\color[HTML]{000000} 0}               & {\color[HTML]{000000} 0}               & {\color[HTML]{000000} 0.0005}          & {\color[HTML]{000000} 0.0116}          \\
\textbf{LSTM-VAE}                                          & {\color[HTML]{330001} 0.4993}          & {\color[HTML]{000000} 0.9929}          & {\color[HTML]{000000} \textbf{0.5556}} & {\color[HTML]{000000} 0.3333}          & {\color[HTML]{343434} 0.3766}          & {\color[HTML]{000000} \textbf{0.9453}}                     & {\color[HTML]{000000} 0}               & {\color[HTML]{000000} 0.2345}          & {\color[HTML]{000000} 0.0253}          & {\color[HTML]{000000} 0.3774}          & {\color[HTML]{000000} 0.6535}          & {\color[HTML]{000000} 0}               & {\color[HTML]{000000} 0.3298}          & {\color[HTML]{000000} 0.0471}          & {\color[HTML]{000000} 0.3770}          \\
\textbf{MSCRED}                                            & {\color[HTML]{330001} 0.3733}          & {\color[HTML]{000000} 0.5890}          & {\color[HTML]{000000} 0.1457}          & {\color[HTML]{000000} 0.3913}          & {\color[HTML]{343434} 0.5261}          & {\color[HTML]{000000} 0.9277}                              & {\color[HTML]{000000} 0.9801}          & {\color[HTML]{000000} \textbf{0.9358}} & {\color[HTML]{000000} 0.2860}          & {\color[HTML]{000000} 0.2836}          & {\color[HTML]{000000} 0.5324}          & {\color[HTML]{000000} \textbf{0.7358}} & {\color[HTML]{000000} 0.2521}          & {\color[HTML]{000000} 0.3304}          & {\color[HTML]{000000} 0.3685}          \\
\textbf{BeatGan}                                           & {\color[HTML]{330001} 0.3742}          & {\color[HTML]{000000} 0.8793}          & {\color[HTML]{000000} 0.2499}          & {\color[HTML]{000000} 0.3224}          & {\color[HTML]{343434} 0.4233}          & {\color[HTML]{000000} 0.9441}                              & {\color[HTML]{000000} 0}               & {\color[HTML]{000000} 0.2390}          & {\color[HTML]{000000} 0.5000}          & {\color[HTML]{000000} 0.1287}          & {\color[HTML]{000000} 0.5360}          & {\color[HTML]{000000} 0}               & {\color[HTML]{000000} 0.2443}          & {\color[HTML]{000000} 0.3920}          & {\color[HTML]{000000} 0.1974}          \\
\textbf{TCN-AE}                                            & {\color[HTML]{330001} 0.3995}          & {\color[HTML]{000000} 0.8166}          & {\color[HTML]{000000} 0.2185}          & {\color[HTML]{000000} 0.1189}          & {\color[HTML]{343434} 0.5795}          & {\color[HTML]{000000} 0.6308}                              & {\color[HTML]{000000} 0}               & {\color[HTML]{000000} 0.2376}          & {\color[HTML]{000000} 0.3258}          & {\color[HTML]{000000} 0.1851}          & {\color[HTML]{000000} 0.4892}          & {\color[HTML]{000000} 0}               & {\color[HTML]{000000} 0.2277}          & {\color[HTML]{000000} 0.1743}          & {\color[HTML]{000000} 0.2806}          \\
\textbf{UAE}                                               & {\color[HTML]{330001} 0.4984}          & {\color[HTML]{000000} 0.8553}          & {\color[HTML]{000000} 0.2961}          & {\color[HTML]{000000} 0.0910}          & {\color[HTML]{000000} \textbf{0.8610}} & {\color[HTML]{000000} 0.6307}                              & {\color[HTML]{000000} 0}               & {\color[HTML]{000000} 0.2418}          & {\color[HTML]{000000} 0.5000}          & {\color[HTML]{000000} 0.1842}          & {\color[HTML]{000000} 0.5568}          & {\color[HTML]{000000} 0}               & {\color[HTML]{000000} 0.2662}          & {\color[HTML]{000000} 0.1540}          & {\color[HTML]{000000} 0.3034}          \\
\textbf{EdgeConvFormer}                                    & {\color[HTML]{330001} 0.5751}          & {\color[HTML]{000000} 0.5759}          & {\color[HTML]{000000} 0.4877}          & {\color[HTML]{000000} \textbf{0.6097}} & {\color[HTML]{343434} 0.4169}          & {\color[HTML]{000000} 0.9229}                              & {\color[HTML]{000000} \textbf{0.9999}} & {\color[HTML]{000000} 0.7565}          & {\color[HTML]{000000} \textbf{0.5343}} & {\color[HTML]{000000} \textbf{0.8022}} & {\color[HTML]{000000} \textbf{0.7086}} & {\color[HTML]{000000} 0.7309}          & {\color[HTML]{000000} \textbf{0.5930}} & {\color[HTML]{000000} \textbf{0.5695}} & {\color[HTML]{000000} \textbf{0.5487}} \\ \bottomrule
\end{tabular}
}
\label{Tab:ExathlonResults}
\end{table*}
\textbf{\textit{Runtime}}
We measured the runtimes for all algorithms on six datasets and normalized the runtimes by the lengths of the respective time series. Fig.~\ref{fig:runtime} shows the average measured runtime of every algorithm for one data point, including training and testing times. BeatGAN, UAE, and TCN-AE run the fastest ($< 10ms$ per data point), EdgeConvFormer, the top-performing model in the evaluation, is the third slowest, while MSCRED took the longest due to the complexity of the algorithm (nearly $60ms$ per data point). Our EdgeConvFormer consumes around 26ms per data point, achieving a good balance between accuracy and efficiency.

\begin{figure}[ht]
\centering
\includegraphics[width=1.0\linewidth]{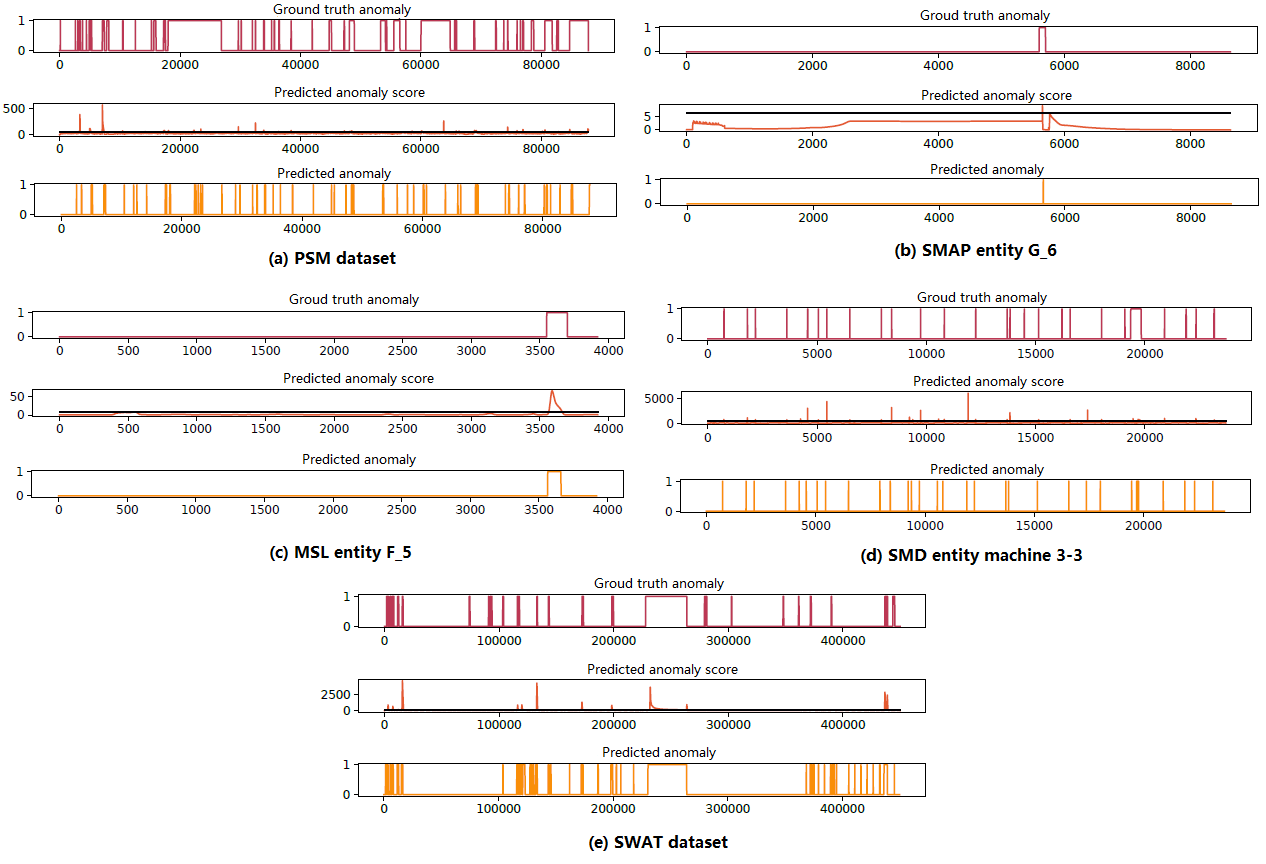}

\caption{Qualitative evaluation of anomaly detection using the proposed EdgeConvFormer model. For each dataset, the wine-red line is the ground truth anomaly, the orange line is the predicted anomaly score, and the yellow line is the predicted anomaly label by applying the best-Fc$_1$ threshold (black line) to the anomaly score.}
\label{fig:result}
\end{figure}

\textbf{\textit{Qualitative evalution}}
Fig.~\ref{fig:result} shows qualitatively the anomaly score and predicted labels of the EdgeConvFormer model with dynamical Gaussian scoring and $Best$-$Fc_1$ threshold on the test set of the five public datasets. An ideal detector would detect at least one time-point in each event (segment) to raise an alarm and has no FPs. Compared with the ground truth labels, the model performs well on each dataset, with lower false positives and missed detection rates. EdgeConvFormer can detect not only point anomalies but also segment anomalies. Although EdgeConvFormer does not detect all points in every anomaly segment, at least one point is detected in almost all anomaly segments.

\begin{figure}[ht]
  \centering
  \begin{subfigure}[b]{0.48\textwidth}
    \includegraphics[width=\textwidth]{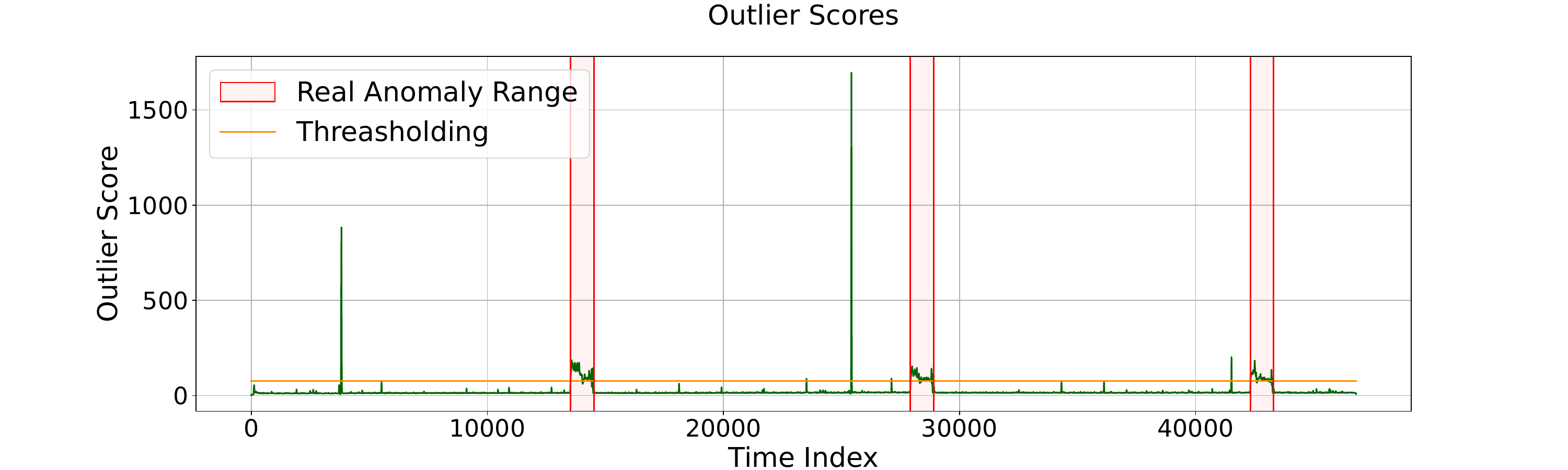}
    \caption{T1 trace of Application 2}
    \label{fig:subfig11}
  \end{subfigure}
  \begin{subfigure}[b]{0.48\textwidth}
    \includegraphics[width=\textwidth]{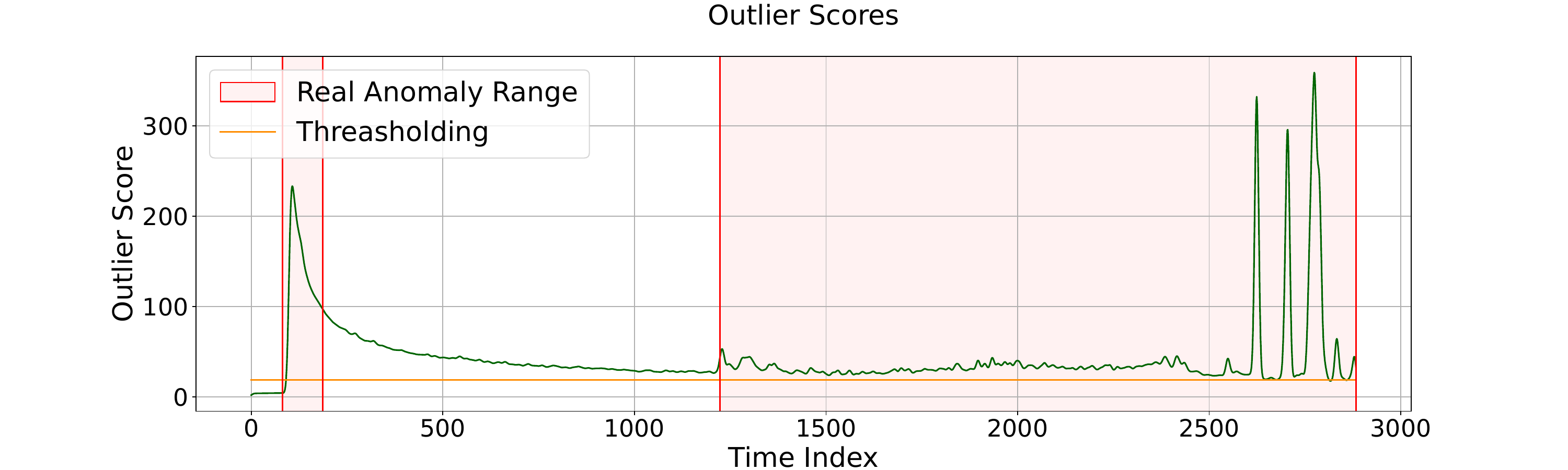}
    \caption{T2 trace of Application 2}
    \label{fig:subfig12}
  \end{subfigure}
  \begin{subfigure}[b]{0.48\textwidth}
    \includegraphics[width=\textwidth]{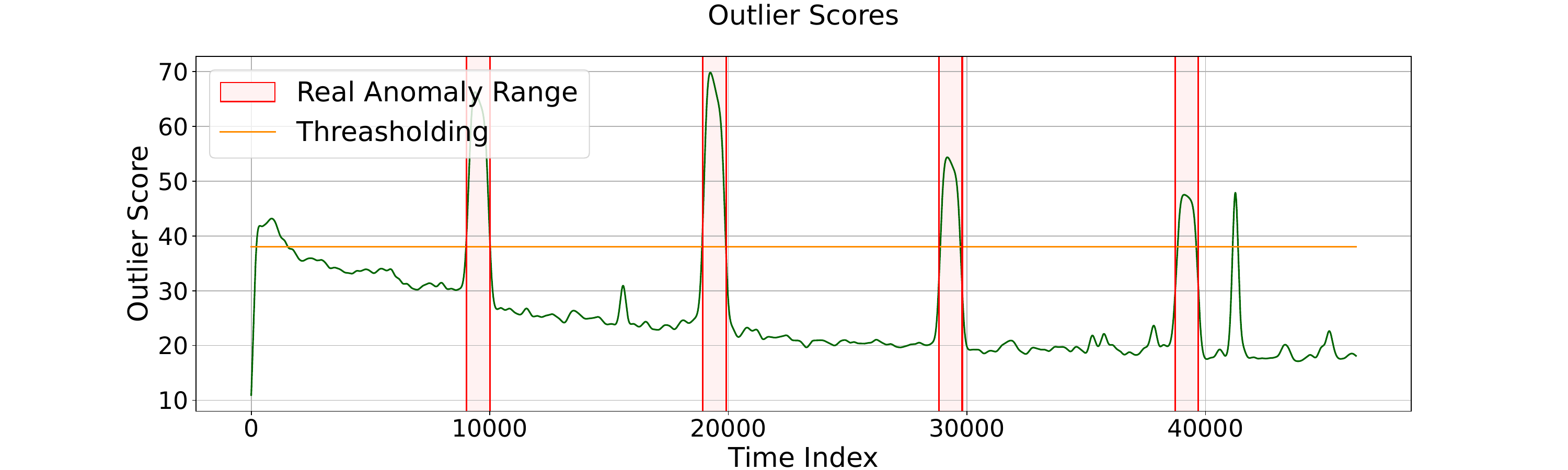}
    \caption{T3 trace of Application 9}
    \label{fig:subfig13}
  \end{subfigure} 
  \begin{subfigure}[b]{0.48\textwidth}
    \includegraphics[width=\textwidth]{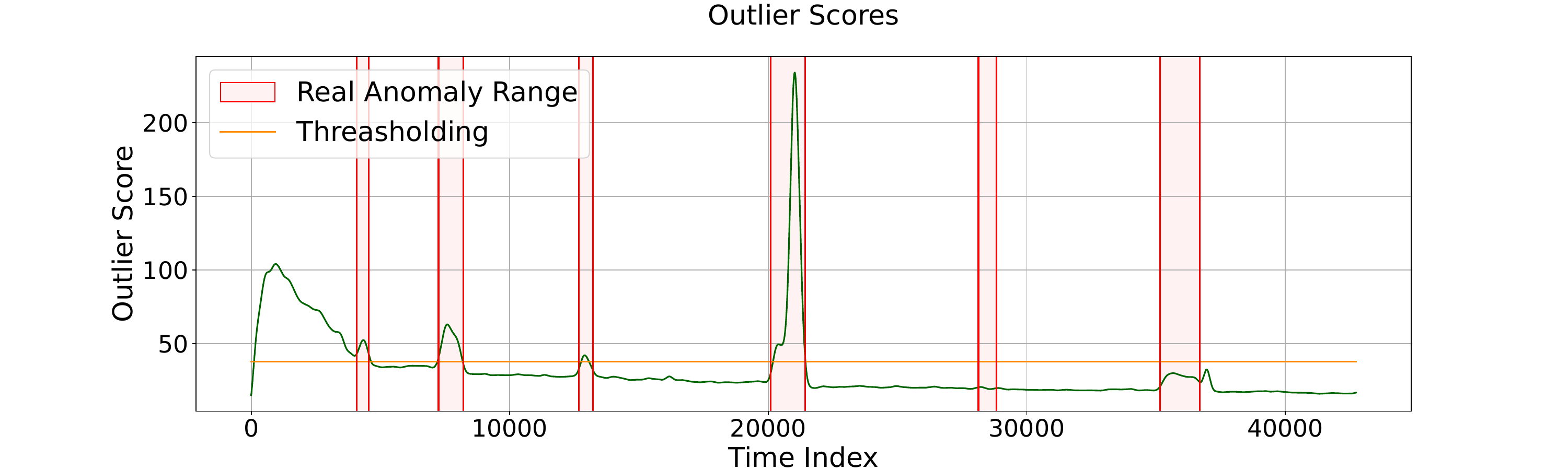}
    \caption{T4 trace of Application 1}
    \label{fig:subfig14}
  \end{subfigure} 
  \begin{subfigure}[b]{0.48\textwidth}
    \includegraphics[width=\textwidth]{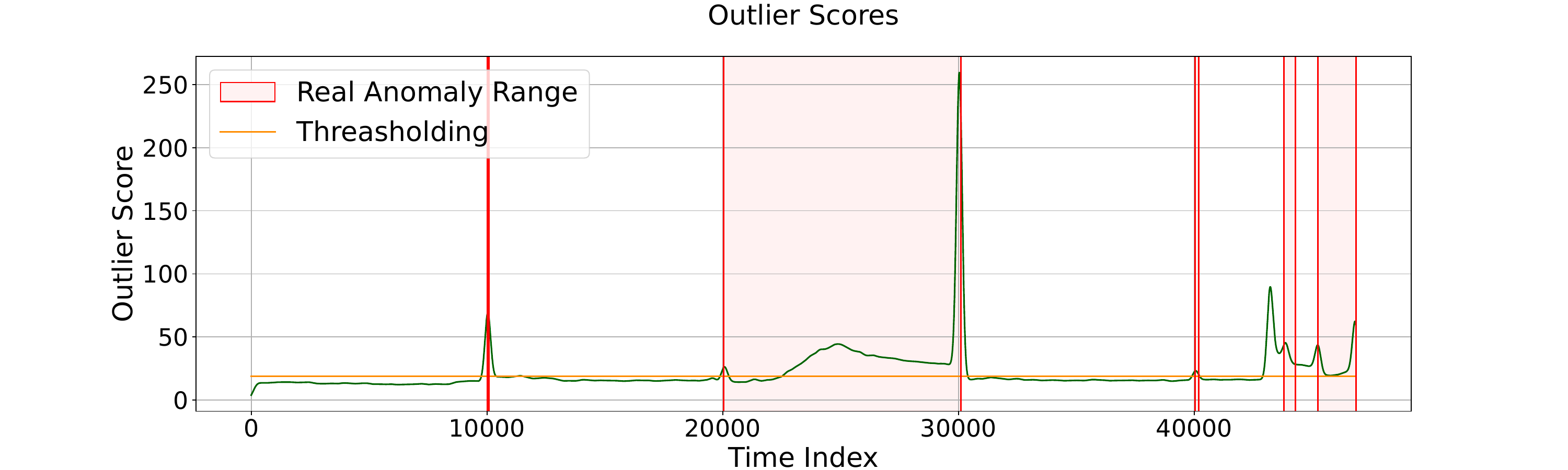}
    \caption{T5 trace of Application 2}
    \label{fig:subfig15}
  \end{subfigure} 
  \caption{Qualitative evaluation of anomaly detection on Exathlon dataset using the proposed EdgeConvFormer model. The pink rectangle is the ground truth anomaly, the green line is the predicted anomaly score (with Gauss\_D\_K scoring), and the yellow line is the tail-p threshold to the anomaly score.}
  \label{fig:exathlon}
\end{figure}

Fig.~\ref{fig:exathlon} shows qualitatively the anomaly score thresholding of the EdgeConvFormer model with the dynamic Gaussian scoring function applied
with gaussian kernel scoring and $Tail$-$p$ threshold on the disturbed traces of the Exathlon datasets. Except for two abnormal segments missed by the T4 trace in application 1 (the second one has a very high anomaly score, but was missed because the threshold was selected too high due to the influence of the high score of the initial false positive), the remaining abnormal segments are all detected, and the detected anomaly range size was consistent with the ground truth.

No matter which evaluation metric is used, the EdgeConvFormer model can achieve the best or comparable results, compared with the SOTA methods. Especially in Exathlon, a high dimension and large size dataset with a strict four-level evaluation mechanism, the EdgeConvFormer model maintains a stable score, indicating that the EdgeConvFormer model has strong abilities to separate normal and abnormal data in the outlier space and to detect anomalies based on threshold selection. It proves that the EdgeConvFormer model is robust and can perform well in anomaly detection of multivariate time series.

\subsection{Ablation studies}
To study the efficacy of Time2Vec embedding, EdgeConv module, and Transformer module specifically in our method, we replace Time2Vec with position encoding in the Transformer, remove the EdgeConv module, and the Transformer module in the four-layer structure of the encoder respectively to observe how the model performs. The results are summarized in TABLE~\ref{Tab:ablation} and provide the following findings. Replacing Time2Vec with position encoding reduces F$_1$, Fpa$_1$, Fc$_1$, AU-ROC, and AU-PRC by 89.71\%, 7.11\%, 19.95\%, 11.68\%, and 26.47\% respectively. It shows that Time2Vec, which is related to the time decomposition technique that encodes a temporal signal into a set of frequencies, can effectively consume temporal information and improve performance. By removing EdgeConv modules the F$_1$ score drops the most (\begin{math}0.3265\rightarrow0.0326\end{math}), and Fpa$_1$, Fc$_1$, AU-ROC, and AU-PRC decrease 7.15\%, 19.64\%, 10.58\%, and 25.07\% respectively, indicating that the EdgeConv module is the most crucial component in representing both the spatial and temporal information. By removing the Transformer module the F$_1$, Fpa$_1$, Fc$_1$, AU-ROC, and AU-PRC decrease 6.00\%, 5.90\%, 8.72\%, 4.86\%, and 13.51\% respectively, not as much as the EdgeConv module, which implies that the EdgeConv extracts the features with the strongest correlation to the center point from the nearest neighbors and it can help Transformer attend to more meaningful information. Nevertheless, using a Transformer to attend to the time dimension and capture the information across long-range timestamps can improve the performance of the model significantly.

These findings suggest that the Time2Vec embedding, the EdgeConv module, and the Transformer module of the proposed EdgeConveFormer model all contribute to its improved anomaly detection ability, which explains its better performance over other state-of-the-art methods.

\begin{table}
\centering
\caption{Ablation of Encoder with the best-F-score threshold on PSM dataset. w/o Time2Vec means replacing Time2Vec with position encoding in the Transformer. w/o EdgeConv and w/o Transformer mean to remove the EdgeConv module and Transformer module in each layer of the encoder, respectively. The best results are highlighted in bold.}

\begin{tabular}{@{}cc|lll|llc|lll|ccc|ccc@{}}
\toprule
\multicolumn{2}{c|}{\textbf{Metric}}                                          & \multicolumn{1}{c|}{\textbf{F$_1$}}                                                                                  & \multicolumn{1}{c|}{\textbf{Fpa$_1$}}                                                                                & \multicolumn{1}{c|}{\textbf{Fc$_1$}}                                                                                 & \multicolumn{1}{c|}{\textbf{AU-ROC}} & \multicolumn{1}{c}{\textbf{AU-PRC}} \\ \cmidrule(r){1-7}

& \textbf{w/o Time2Vec}            	&\multicolumn{1}{c|}{0.0336} 	
                               	&\multicolumn{1}{c|}{0.9047} 	
                               	&\multicolumn{1}{c|}{0.6007} 	
                               & \multicolumn{1}{c|}{0.6037}          & \multicolumn{1}{c}{0.3795}          \\

  & \textbf{w/o EdgeConv}            	&\multicolumn{1}{c|}{0.0326} 	
                               	&\multicolumn{1}{c|}{0.9044} 	
                               	&\multicolumn{1}{c|}{0.6030} 	
                               & \multicolumn{1}{c|}{0.6112}          & \multicolumn{1}{c}{0.3867}          \\

& \textbf{w/o Transformer}             	&\multicolumn{1}{c|}{0.3069} 	
                                    &\multicolumn{1}{c|}{0.9165}                                     	&\multicolumn{1}{c|}{0.6850} 	
                               & \multicolumn{1}{c|}{0.6503}          & \multicolumn{1}{c}{0.4464}          \\

 & \textbf{EdgeConvFormer}  & \multicolumn{1}{c|}{\textbf{0.3265}}                                          & \multicolumn{1}{c|}{\textbf{0.9740}}            & \multicolumn{1}{c|}{\textbf{0.7504}}                      & \multicolumn{1}{c|}{\textbf{0.6835}} & \multicolumn{1}{c}{\textbf{0.5161}}          \\ \bottomrule
\end{tabular}

\label{Tab:ablation}

\end{table}

\section{Conclusion}
This paper proposes an EdgeConvFormer model that combines Time2Vec, dynamic graph CNNs (EdgeConv), and Transformers for anomaly detection in multivariate time series. Time2Vec encodes input embeddings to obtain position information and periodic/non-periodic behavior of time series simultaneously. The multi-layered architecture of stacking EdgeConv and Transformers can alternately and progressively refine spatiotemporal features to detect anomalies in multivariate time series data more accurately. The EdgeConvFormer model achieves state-of-the-art results and robustness under different evaluation metrics for extensive real-world datasets. Future work can consider changing the point-wise representation in graph CNN and self-attention to a sub-series level to reduce computational complexity and memory consumption.

\section*{Acknowledgement}
The authors would like to acknowledge the support from Australia Research Council Discovery Project DP210103631: AI Assisted Probabilistic Structural Health Monitoring with Uncertain Data.

\bibliographystyle{model1-num-names}

\bibliography{cas-refs}


\end{document}